\documentclass[runningheads]{llncs}
\usepackage{graphicx}
\usepackage[utf8]{inputenc}

\usepackage{amsmath}
    \usepackage{bbm}
\usepackage{amssymb}
\usepackage[table]{xcolor}

\usepackage{amsthm}
\usepackage{mathtools}
\usepackage{soul}
\usepackage{url}  
\usepackage{breakurl}  
\usepackage{bm}
\usepackage{stmaryrd}

\usepackage[ruled]{algorithm2e}

\usepackage{subcaption}
\usepackage{hyperref}
\usepackage{enumitem}
\usepackage{booktabs} 
\usepackage{multirow} 
\usepackage{siunitx} 
\usepackage{amsfonts} 
\usepackage{xfrac}
\usepackage{mathtools}
\usepackage{wrapfig}
\usepackage[skip=0.333\baselineskip]{subcaption}
\usepackage{stackengine}
\newcommand{\NA}{{--}} 
\usepackage[numbers]{natbib}
\usepackage{multirow}
\newcommand{\argmax}[1]{\underset{#1}{\operatorname{arg}\,\operatorname{max}}\;}
\usepackage{ulem}

\usepackage{bigfoot}
\DeclareNewFootnote[para]{default}

\usepackage{todonotes}
\usepackage{soul}
\usepackage{natbib}
\usepackage{todonotes}
\usepackage{comment}

\begin{document}
\title{Interpretability in Safety-Critical Financial Trading Systems}
\author{Anonymous Authors}
\author{Gabriel Deza\inst{1} \and Adelin Travers\inst{1} \and Colin Rowat\inst{2} \and Nicolas Papernot\inst{1}}
\authorrunning{F. Author et al.}
\institute{University of Toronto and Vector Institute, Toronto, Canada \and University of Birmingham, Birmingham, England}
\maketitle              %
\begin{abstract}
Sophisticated machine learning (ML) models to inform trading in the financial sector create problems of interpretability and risk management. Seemingly robust forecasting models may behave erroneously in out of distribution settings. In 2020, some of the world's most sophisticated quant hedge funds suffered losses as their ML models were first underhedged, and then overcompensated.

We implement a gradient-based approach for precisely stress-testing how a trading model's forecasts can be manipulated, and their effects on downstream tasks at the trading execution level. We construct inputs -- whether in changes to sentiment or market variables -- that efficiently affect changes in the return distribution. In an industry-standard trading pipeline, we perturb model inputs for eight S\&P 500 stocks. We find our approach discovers seemingly in-sample input settings that result in large negative shifts in return distributions.

We provide the financial community with mechanisms to interpret ML forecasts in trading systems. For the security community, we provide a compelling application where studying ML robustness necessitates that one capture an end-to-end system's performance rather than study a ML model in isolation. Indeed, we show in our evaluation that errors in the forecasting model's predictions alone are not sufficient for trading decisions made based on these forecasts to yield a negative return.
\keywords{ML Interpretability \and Financial Trading \and Risk Management}
\end{abstract}
\section{Introduction}
\label{sec:introduction}
Deep Neural Networks (DNNs) have been proposed in finance for over three decades~\cite{FinML,trippi1992neural} due their ability to learn more complex and non-linear mappings compared to classical time series models. However, the additional complexity of DNNs makes their interpretability of particular importance when it comes to widely adopting them in production settings. We study financial trading systems built on such DNNs in order to develop a method for better interpretability. {\let\thefootnote\relax\footnote{{All correspondance to gabe.deza@mail.utoronto.ca}}}

Modern financial regulations (\textit{e.g.} Basel III~\cite{noauthor_basel_nodate}) require stress testing models, including 
on past catastrophic events. %
For instance, regulators may be interested in learning the worst-case outcomes of a model given a scenario of adverse market conditions~\cite{mcneil2015quantitative}. Existing approaches for interpretability of DNNs are not directly applicable to this setting for two reasons. First, time series forecasting models used in finance output distributions rather than single-point predictions. Second, these distributions predicted by the model are not the outcome of the pipeline; instead, they are used as inputs to inform trading decisions.

To address the specificities of safety-critical financial trading systems, we propose an interpretability and risk estimation method to synthesize adverse market conditions.  
These synthetic market conditions reveal to a financial institution the factors which influence the trading decisions made by its systems. 
In particular, our method shows how seemingly regular market conditions can be manipulated to introduce market adversity undermining a DNN's forecast. In an application to an industry-standard trading pipeline, we for instance show how our method provides insights on feature importance, choice of trading signals and the pipeline's robustness.
To summarize, our contributions are as follows:

\begin{itemize}
    \vspace{-5pt}
    \item We instantiate a gradient-based optimization method that informs model owners on the sensitivity of their forecasting model to its input features. %
    \item We develop an industry-standard daily stock trading pipeline. Our pipeline integrates Twitter sentiment analysis together with several trading strategies and achieves non-trivial performance and returns across 8 S\&P 500 tickers under realistic conditions. We open-source our pipeline (see Section~\ref{ssec:abstract_view}). %
    \item We demonstrate on our trading pipeline that our method synthesizes adverse market conditions that illustrate the influence of different model inputs on the trading pipeline's returns. For instance, increasing adversity of the market conditions results in decreasing returns by almost $\sfrac{1}{3}$ (21.3\% to 7.7\%) whereas decreasing adversity yields a $5\times$ larger return (2.2\% to 10.3\%).
    \item Our method provides controllability on the adversity (or lack of adversity) of the market conditions synthesized via the perturbation amount $\epsilon$ introduced to each feature used to model market conditions. We show that varying $\epsilon$ can shift and change the distribution of returns, providing an intuitive means to understand how each of these features influence trading decisions. %
    \item Our method exposes how the mean of the output distribution from our forecasting model is easily manipulated (requiring a smaller total perturbation $\epsilon$) compared to other parameters of the distribution (such as the confidence). Iterating on trading strategies that rely less on such non-robust parameters directly can help make pipelines more robust to such adversity.

\vspace{-5pt}
\end{itemize}

\section{Background}
\vspace{-5pt}
\subsection{Trading Pipelines for Stock Forecasting}
\label{ssec:stock forecasting}

A company may issue stock also called shares, units of ownership of the company, which are traded (bought or sold) on exchanges. A \textbf{ticker} is  used to designate the stock on the exchange and look up price change information.  We work with tickers from the \textbf{S\&P 500}, the largest 500 companies traded in the U.S. In first approximation, a stock may only be traded during an exchange's \textbf{market hours}. (For example, the New York Stock Exchange's opens to trading on weekdays at 9:30 AM and closes at 4:00 PM.) For each trading day, a ticker's \textbf{open and close price} is the price quoted at market open and market close, respectively.

We consider a financial pipeline that tackles the problem of forecasting, \textit{i.e.}, tries to predict the price changes ahead of time. Formally, after collecting daily open and close prices and potential price-change factors, \textit{e.g.}, sentiment data, over $k_{\text{past}}$ days for a given ticker, a forecast generates a prediction of the daily open and close prices for the next $k_{\text{future}}$ days of trading. This forecast is then used to make trading decisions. %

\vspace{-5pt}
\paragraph{General Trading Pipeline Formulation.}
\label{ssec:abstract_view}
\vspace{-5pt}

The entire trading process, and resulting algorithmic systems, can be abstracted into 3 major steps described below. %

\begin{itemize}
    \vspace{-5pt}
    \item \textbf{Data Collection:} Common data includes historical and/or real time financial data provided by exchanges, sentiment data directly collected from news sources or provided by data vendors such as Bloomberg or Reuters~\cite{noauthor_finding_2017, noauthor_thomson_nodate}. 
    \item \textbf{Forecasting}: Collected data is then analyzed for useful trading signals. This analysis is performed using a model $\mathcal{M}$, whose complexity can range from a linear model or basic pattern analysis to Deep Neural Networks (DNNs). %
    \item \textbf{Trading Decisions:} Using the forecasts from the previous step, a trading decision is made to buy or sell $V$ shares at price $P$.
    \vspace{-5pt}
\end{itemize}

Obtaining a realistic test bench encompassing all of these steps is complex and time consuming. To aid future work in this setting, we include the code to reproduce our pipeline and the results within our work.\footnote{Our code: \burl{https://anonymous.4open.science/r/FinancialML_Interpretability-68F4}} \vspace{-5pt} %

\subsection{Deep Neural Networks for Probabilistic Forecasting\vspace{-4pt}}
\label{ssec:DNN for forecasting}

Finance is a data-rich domain~\cite{ko-kr-17} making methods that benefit from this information, such as DNNs, attractive. This potential has brought significant interest from the financial and ML academic communities for  finance and trading tasks like forecasting or portfolio management\cite{https://doi.org/10.1002/asmb.2209,FinML,8324237,chen2021deep}. In addition, large bank and investment firms have recently invested heavily in such research \cite{rbc-aiden,ko-kr-17}. 

When choosing a forecast model, it is possible to output either a single data value (point forecast)~\cite{KRAUSS2017689} or a probability distribution~\cite{pmlr-v119-dang-nhu20a}. Probabilistic forecasting is preferred over point forecasting as confidence scores or risk estimates can be derived from the forecasted distribution~\cite{fraccaro2016sequential}. %
While the forecast model specifically outputs the parameters generating the distribution rather than the distribution itself, for simplicity we henceforth conflate both.
Architectures for time series forecasting include recurrent neural networks (RNNs)~\citep{salinas2020deepar, smyl2020hybrid}, convolutional neural networks~\citep{oord2016wavenet,borovykh2017conditional,bai2018empirical} with $1$-dimensional convolutions over time, as well as transformers~\citep{vaswani2017attention,lim2019temporal,li2019enhancing} using attention-based mechanisms. In our work, we leverage RNNs for probabilistic forecasting, described in Section~\ref{sssec:probabilistic forecasting}, because of their superior performance in temporal settings like forecasting.

Nonetheless, these proof-of-concept research advances may be difficult to readily deploy in production settings as current DNNs lack proper input-output relationship transparency. We address this limitation with our work. %

\subsection{Interpretability in ML}
\label{ssec:interpretability in ML}
\vspace{-4pt}
In select domains like computer vision and natural language processing, interpretability of ML models is well studied%
~\cite{zhang2021survey}. In vision, techniques visualize how a prediction stems from different input pixels~\cite{simonyan2013deep}, the learned weights of a model~\cite{nguyen2016synthesizing}, or its receptive field~\cite{zhou2014object}. To the best of our knowledge, leveraging gradients for interpretability in domains like finance remains unexplored. Financial systems raise several challenges. First and foremost, DNNs only inform  trading strategies; this calls for an end-to-end outlook over the entire pipeline. In addition, financial data is inherently more difficult for humans to interpret when compared to computer vision and NLP.
\vspace{-5pt}
\section{Gradient-based Interpretability for Forecasting Models}
\label{sec:gradient_algorithm}
\vspace{-5pt}
In this section, we introduce the proposed interpretability method with a formal end-to-end pipeline for financial trading. %
We leverage gradient information of model $\mathcal{M}$ and characteristics of the entire trading pipeline to  understand edge case model behaviour. %
While introduced in the financial trading domain, our method is not limited to it; it is applicable broadly to forecasting problems. \vspace{-5pt}%

\subsection{Preliminary}
\vspace{-5pt}
Formally, let $z_{1:T}$ be a single target time series of length $T$ and $\mathbf{X}$ a set of associated covariates series $\mathbf{X}_{1:N, 1:T} \triangleq \{\bm{x}_{i_,1:T}\}_{i=1}^{N}$. A (DNN-based) forecast is a generated probability distribution over future (unobserved) values $z_{T+1:T+\tau}$ of length $\tau$ conditioned on past time series values $z_{1:T}$ and covariates $\mathbf{X}_{1:N,T:T+\tau}$ %
using a neural network $\mathcal{M}$ parameterized by $\Phi$, the model's weights and biases. %
\begin{equation}
\label{eq:1}
    \Pr(z_{T+1:T+\tau}\ |\ z_{1:T}, \mathbf{X}_{1:N,1:T}) = \mathcal{M}_\Phi(z_{1:T}, \mathbf{X}_{1:N,1:T})
\end{equation}
It is common to follow the Markov assumption in forecasting where the next $\tau$ days is likely not a function of the entire history length available $T$ but rather a smaller portion that represents the recent past. We only use the past $k$ temporal observations of $z$ and $\mathbf{X}$ for the next $\tau$ days forecast. The estimated probability distribution is defined by its $n$ distribution parameters. Since the output of the model is $\tau$ distributions, each defined by $n$ parameters, we denote the entire output as  $\mathbf{\Theta}_{1:n,1:\tau} \triangleq \{\bm{\theta}_{i_,1:\tau}\}_{i=1}^{n}$; this can be conceptualized as a $n \times \tau$ matrix. Hence, Equation~\ref{eq:1} can be rewritten as Equation~\ref{eq:forecast_fixed}. %
\begin{equation}
\mathbf{\Theta}_{1:n,T+1:T+\tau} = \mathcal{M}_\Phi(z_{T-k:T}, \mathbf{X}_{1:N,T-k:T})
\label{eq:forecast_fixed}
\end{equation}

Definitions of interpretability vary with the goals of model owners and the settings models are deployed in. In financial trading pipelines robustness is primordial, and thus we set out the following requirements for our interpretability method:
\begin{itemize}
    \item [G1:] Understanding of how a pipeline's outputs are going to behave in the face of deviations from their expected inputs. For instance, given an input $x$ based on average historical data and a model prediction $\mathcal{M}(x) = y$, we want to understand $\mathcal{M}(x + \epsilon)$ for $\epsilon$ capturing deviations from this historical data. Note that we need to take care to upper bound $\epsilon$ to ensure that we consider edge cases that remain realistic (i.e., that could be realized). 
    \item [G2:] Developing intuition on how the pipeline will react to an unknown input (i.e., For some regular input $x$ develop intuition for $\mathcal{M}(x)$).
    \item [G3:] Developing intuition on what the pipeline input must have been to achieve a specific output (i.e., For some output $y$ develop intuition on $\{x | \mathcal{M}(x) = y\}$).\vspace{-5pt}
\end{itemize}
\subsection{Our algorithm for model interpretability}
\vspace{-5pt}
At a high level, our method uses model gradients to synthesize adverse settings by manipulating the covariate features $\mathbf{X}$ resulting in forged features $\mathbf{\hat{X}}$ such that the original output parameters $\mathbf{\Theta}$ change in a specific direction and amount. The resulting changed output parameters are denoted $\mathbf{\hat{\Theta}}$. These forged features $\mathbf{\hat{X}}$ and the resulting outputs $\mathbf{\hat{\Theta}}$ both provide interpretability of the model by quantifying its sensitivity to changes to the inputs of the forecasting problem. 

Specifically, given a direction (up ($\uparrow$) or down ($\downarrow$)) and a distribution parameter index $p$ is selected, the model inputs are perturbed from $\mathbf{X}$ to $\mathbf{\hat{X}}$ such that the model outputs $\mathbf{\Theta}_{p,T+1:T+\tau}$ are manipulated in that given direction, resulting into $\mathbf{\hat{\Theta}}_{p,T+1:T+\tau}$. We accomplish this by leveraging partial derivative computation through the model $\mathcal{M}$. For instance, the partial derivative of the output distribution parameter $\mathbf{\Theta}_{i,t}$ with respect to the input $\mathbf{X}_{j,l}$ captures the sensitivity of the $i^{th}$ distribution parameter at time $t$ to the $j^{th}$ input feature at time $l$. Our work is based on two (independently, well known) facts: %
(i) partial derivatives provide a wealth of information on a model's behaviour and (ii) a model owner has full access to model parameters and can thus explicitly compute any partial derivative with regard to any input they wish.
Together this means that a model owner can explicitly explore the entire range of model behaviours for risk analysis by leveraging the information contained in the gradients.
\setlength{\textfloatsep}{3pt}
\begin{algorithm}[t]
\caption{Interpretability of ML Forecasting Models Algorithm}
\scriptsize
\SetAlgoLined
\LinesNumbered
\SetKwInOut{Input}{input}
\SetKwInOut{Output}{output}
\SetKw{KwBy}{by}
\Input{Index of Distribution parameter $p \in [1,n]$, direction to perturb distribution parameter $d \in \{\uparrow,\downarrow\}$, Forecasting length $\tau$, Historical length $k$, historical covariate features $\mathbf{X}_{1:N,T-k:T}$, historical target values $z_{T-k:T}$, perturbation amount $\epsilon$ and number of iterations $R$}
\Output{$\mathbf{\hat{X}}_{1:N,T-k:T}$}
\BlankLine
\tcc{Initially start with benign features}
$\mathbf{\hat{X}}_{1:N,T-k:T} = \mathbf{X}_{1:N,T-k:T}$ \\
\tcc{Iterate over number of perturbations}
\For{$j\gets1$\KwTo $R$ \KwBy $1$}{
    \tcc{Iterate over the prediction length}
    \For{$t\gets T+1$ \KwTo $T+\tau$}{
        \tcc{Iterate over the historical length} 
        \For{$s\gets T-k$ \KwTo $T$}{
            \tcc{Find input with largest gradient magnitude}
            $i^{*} = \argmax{i} |\frac{\partial \mathbf{\hat{\Theta}}_{p,t}}{\partial \mathbf{\hat{X}}_{i,s}}| $\\
            \tcc{Check bound characteristic of feature $i^{*}$}
            \uIf{\texttt{Checkbounds}($i^{*}$)}{
            $\mathbf{\hat{X}}_{i^{*},s} = \mathbf{\hat{X}}_{i^{*},s} +\Big(d \times \texttt{sgn}(\frac{\partial \mathbf{\hat{\Theta}}_{p,t}}{\partial \mathbf{\hat{X}}_{i^{*},s}})\Big) \times \epsilon$}
            \Else{
            \tcc{Jump to line 8 but do not consider the argmax over $i = i^{*}$}
            }
        }
    }
}
\label{algo1}
\end{algorithm}

In Algorithm~\ref{algo1}, the features are perturbed as long as they stay within their historical range (recall our requirement G1). This is checked by a function which we refer to as $\texttt{Checkbounds}$; it serves as a simple non-ML heuristic for input anomaly detection, likely on top of the already implemented security checks inherent to any realistic pipeline. Such anomalies are avoided as our interpretability algorithm focuses on understanding how models behave in their most common and close to valid setting (G1, G2, G3) and not in settings that are obviously out of distribution. Otherwise, the feature with the second largest gradient is selected until a feature that satisfies this criterion is chosen.\footnote{Albeit this case did not occur in our experiments, if all features cannot be perturbed such that they stay within the historical bounds, the algorithm terminates early.}

Next, we outline two key design choices made for Algorithm~\ref{algo1}. Together, they address the lack of interpretability of ML models according to G1, G2, G3. \vspace{-5pt}

\paragraph{Gradients.}
\vspace{-5pt}
Gradients offer rich information about the direction in which features need to be perturbed to achieve a desired change in outputs (G1). Specifically, we use the sign and magnitude of the partial derivative  $\frac{\partial y}{\partial x}$ to decide on the direction and magnitude of the perturbation. 
As Algorithm~\ref{algo1} only perturbs the features with the largest gradients, the resulting perturbed features characterize which features best explain a model's reaction to unknown inputs (G2 and G3).
\paragraph{Perturbing benign features.}\vspace{-5pt} Algorithm~\ref{algo1} returns the original features perturbed minimally to achieve the intended change in model output. %
By perturbing from historical market conditions, we obtain conditions that seem like regular inputs to model owners yet can exhibit irregular output behaviour. Observe that this is similar to how mutation-based fuzzing\cite{bohme2017coverage} modifies valid inputs to identify edge case vulnerabilities rather than resorting to random input generation.\vspace{-5pt}

\subsection{Exploring model behaviour with algorithm hyper-parameters}
Algorithm~\ref{algo1} accepts several hyper-parameter to allow for model owners to explore a specific section of the range of possible model behaviours. 
\paragraph{Parameter and direction.}\vspace{-5pt} A practitioner can simulate different trading scenarios by controlling which \textit{direction} an \textit{output distribution parameter} is modified in. For instance, increasing the standard deviation implies lower confidence in forecasts and hence higher risk if a trade is performed based on such forecasts. Such a scenario allows model owners to (1) understand what are the features affecting model confidence and (2) investigate the robustness of trading strategies on top of low confidence forecasts (see Section~\ref{ssec:pipeline-interpretability}). In a similar fashion, the mean of the forecasts represents the expected value of the returns. If we increase the mean of the learned distributions, the inputs are perturbed to reflect synthetic settings where the model is more likely to perform a trade.

\paragraph{Perturbation amount.}\vspace{-5pt}
Algorithm~\ref{algo1} is run for $R$ iterations where the features are perturbed by a small value $\epsilon$ at each iteration. Both hyperparameters allow model owners to tune the adversity of the synthesized setting as they control the how much the perturbed input deviates from historical data for the financial pipeline's inputs. Small perturbations can help diagnose model sensitivity to specific edge cases that give a worse case in settings that are closer to historically regular data; this is particularly true if ``small'' perturbations result in a large impact on the model output. A large $\epsilon$ can model the worst case scenario achievable by exposing the bounds of the distribution parameters that the model can achieve. Such perturbations simulate the worst case when the input features are outside their historical range. This is not entirely a hypothetical scenario: for instance, events 25 standard deviations away from the mean prediction were observed by major financial industry players during the 2007 financial crisis~\cite{noauthor_goldman_2007}. %
\vspace{-10pt}
\subsection{Pipeline interpretability}
\label{ssec:pipeline-interpretability}
\vspace{-5pt}
The forecasting model represents only a fraction of an end-to-end trading pipeline (recall Section~\ref{ssec:abstract_view}). 
In practical deployments, trading strategies use the outputted forecast distribution parameters for profit generation. 
A number of related work have considered the effect of pre-processing on gradient-based perturbations of ML models~\cite{kurakin2016adversarial,carlini2018audio} but, to the best of our knowledge, we are \textit{the first to consider the effects of post-processing} on a model's robustness and interpretability. In our case, this means that a  change in a distribution parameter might not result in any change in the trading strategy depending on its complexity. We thus now consider these trading strategies to ensure our interpretability algorithm is able to capture the end-to-end pipeline's behaviour.

Here, we concentrate on algorithmic trading strategies for which the trading decision is automated---in its simplest form, rule based decisions where a trade occurs if a condition is met, \textit{e.g.}, the forecast  price passing a threshold which can be modeled as a step function. We make this decision for two reasons (i) human traders are known to correct the asset price of their models~\cite{volatility_smiles} and (ii) for risk estimation we require a large number of simulations which is only feasible algorithmically. For instance, a simple trading strategy might be that the forecasted mean must be positive to initiate a trade (buying and later selling). If the mean is originally 4\% and decreased 2\% due to Algorithm~\ref{algo1}, although the mean has indeed changed, the trade  still occurs. 

To specifically aim for a change in the profits of a trading strategy, Algorithm~\ref{algo1} can be modified to take the gradient through the return of the trading strategy as well. In such a case, a non-smooth trading function (e.g, a thresholding strategy in Section~\ref{ssec:trading strats})  results into a uninformative gradient.In our case, the thresholding trading strategies are a function of the mean and the standard deviation of the distribution. We manipulate both of these distribution parameters such that the threshold is met (or no longer met) which results into certain trades occurring or no longer occurring, ultimately changing the return of the strategy.

\section{A Sentiment-based Stock Forecasting and Trading Pipeline}
\vspace{-5pt}
To benchmark our approach for interpretability, we first implement a working industry-standard financial pipeline in Section~\ref{ssec:trading system} along with methods and metrics to evaluate its effectiveness in Section~\ref{ssec:eval setup}. %
We note that this is a significant contribution in itself given that such pipelines are often proprietary and little details can be found in the public domain. %
\vspace{-5pt}
\subsection{Trading System}
\label{ssec:trading system}
\begin{figure}[t]
\begin{center}
\includegraphics[width=0.7\columnwidth]{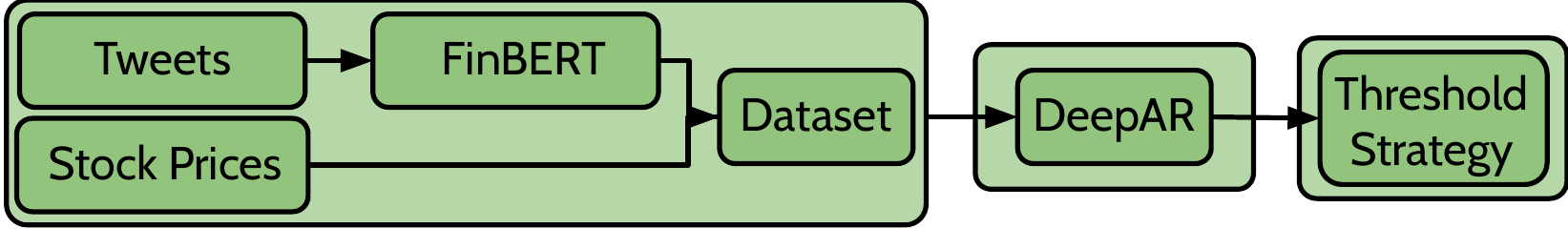}
\end{center}
\vspace*{-7mm}
\caption{Diagram of trading pipeline implemented.}%
\label{fig:pipeline}
\end{figure}

\subsubsection{Data Collection.}
\label{ssec:data}
To avoid selection bias and test generalizability, we randomly select 8 S\&P 500 tickers shown in Table~\ref{table:companies} along with relevant ticker information. In the remainder of this manuscript. we use ticker and company names interchangeably to refer to the companies stock and price over time.

\begin{table}[t]
\resizebox{\linewidth}{!}{
\begin{tabular}{ccccc}
\toprule
\textbf{Ticker} & \textbf{Company} & \textbf{Sector} & \textbf{Market Cap (B \$)} & \textbf{Volume of Tweets (K)} \\
\midrule
HAS & Hasbro & Leisure Products & 13.15 & 1200 \\
ADSK & AutoDesk & Information Technology & 60.40 & 764 \\
XLNX & Xilinx & Semiconductors & 30.93 & 84.7 \\
CAH & Cardinal Health & Health Care Distributors & 16.35 & 45.6 \\
BWA & BorgWarner & Auto Parts \& Equipment & 13.02 &37.9 \\
CHTR & Charter Communications & Cable \& Satellite & 145.45 & 28.1 \\
CE & Celanese & Specialty Chemicals & 18.82 & 24.7 \\
FANG & DiamondBack Energy & Oil \& Gas Exploration \& Production & 15.55 & 16.5 \\
\bottomrule
\end{tabular}
}
\caption{Information on the 8 randomly selected S\&P 500 tickers considered.} %
\label{table:companies}
\end{table}

\begin{itemize}
    \vspace*{-3mm}
    \item \textbf{Tweets:} For each ticker in Table~\ref{table:companies}, we collect tweets from January 1st 2016 to January 1st 2021 (5 year period). Collected tweets are searched to either contain the company name or ticker name of each company.
    \item \textbf{Prices:} The open and close price for each ticker was collected for the same period as the tweets via the Yahoo Finance API. Prices were collected at a daily frequency for the past 5 years. At each interval, there is a quote for the open price and close price. In addition, price data is collected at a hourly frequency for the last 1.5 years. As prices are non-stationary (changing mean and variance over time), we forecast the log difference of each day's close price and open price $log(\frac{r_{\text{close}}}{r_{\text{open}}})$ to get a stationary time series.
    \item \textbf{Sentiment Analysis:} We extract sentiment scores for each tweet which will be used as inputs to our forecasting model. We use FinBERT, a BERT model trained on a financial sentiment dataset~\citep{araci2019finbert}. Given a tweet, FinBERT outputs 3 scores: (1) Positive (2) Neutral and (3) Negative sentiment score. Applying the softmax function we get a probability of a tweet having that sentiment. To forecast the log difference in open and close price for day $T$, we aggregate 13 features (see Table~\ref{table:sentiment features} in the Appendix) derived from tweets up to 24 hours before 9:30 AM on day. Monday's uses the past 72 hours to make use of tweets that occurred over the weekend while the markets were not open to trade.
\end{itemize}

\subsubsection{Forecasting.}
\label{sssec:probabilistic forecasting}
\vspace*{-6mm}
As mentioned in Section~\ref{sec:gradient_algorithm}, probabilistic forecasting is interested in forecasting the $\tau$ distributions, denoted $\mathbf{\Theta}_{1:n,T+1:T+\tau}$:
\begin{equation*}
\mathbf{\Theta}_{1:n,T+1:T+\tau} = \mathcal{M}_\Phi(z_{T-k:T}, \mathbf{X}_{1:N,T-k:T})
\end{equation*}
In our pipeline, we train a model $\mathcal{M}$ with input covariates $\mathbf{X}_{1:N,T-k:T}$ as the 13 features derived from Twitter and the target time series $z_{T-k:T}$ as the log difference in price. We use the DeepAR~\citep{salinas2020deepar} architecture (see Figure~\ref{fig:deepar_arch} in Appendix), an autoregressive RNN tailored for time series forecasting in the univariate setting. RNNs are particularly well suited for this task because they keep an internal state that allows them to output forecasts of variable length. Here, autoregressive refers to the forecasting model using past observations of its input to forecast future behaviour.  %
We use the Python implementation of this model available in the GluonTS~\citep{alexandrov2019gluonts} library for our experiments. DeepAR allow for flexible parametric and non-parametric output distributions by using a projection layer to map the RNN output to parameters defining the chosen distribution function. %
 
\subsubsection{Trading Decisions}
\label{ssec:trading strats}
\vspace*{-6mm}
The forecasting models output $\tau$ distributions. All distribution parameters are in the log difference domain and hence we iteratively undo the log difference to get a forecasted distribution in the return space.

As we consider daily trading, we start the trading day entirely in cash and end the day entirely in cash. Assuming we must buy a stock before selling it (i.e., only taking long positions), we only trade when the close price is greater than the open price. %
At a high level, a trading strategy  involves two decisions: (1) When do we take an action (buy or sell)? (2) How much to invest per trade?

\paragraph{When do we trade?}
\vspace*{-4mm}
We consider a simple thresholding strategy as defined: If the predicted difference $\hat{y} \geq \tau$, for $\tau \geq 0$, we buy shares at 9:30 AM at the open price and sell them back at 4:00 PM at the close price for a return of $y\%$.

We work with two threshold values,$\tau = 0$ and $\tau = \mu_{y} + \sigma_{y}$, where $\mu_{y}$ and $\sigma_y$ refer to the mean and standard deviation of the ground truth returns over the past $k$ days, respectively. The $\tau = 0$ strategy trades on any positive signal while the $\tau = \mu_{y} + \sigma_{y}$ strategy is more prudent, requiring the returns to be one standard deviation above those of the past $k$ days.

\paragraph{How much to invest?}
\label{sssec:how_much_trade}
\vspace*{-4mm}
The Kelly criterion~\cite{kelly_fraction} is often used in financial mathematics when determining how much to invest. For each trading opportunity, we compute the Kelly fraction $f$ which represents what fraction should be invested depending on the expected return. The Kelly fraction is calculated as shown in Equation~\ref{eq:kelly} where the win percentage  $W$ is the percentage of trades that resulted into a profit and $R$ is the ratio of positive returns to negative returns (sometimes referred to as the ratio between historical gains and losses). The Kelly fraction allows trading proportional to the strength of the signal from the forecast. 
\begin{equation}
\label{eq:kelly}
f = W - \frac{1-W}{R}
\end{equation}

\subsection{Evaluation Setup}
\label{ssec:eval setup}
We divide our evaluation metrics into error, accuracy and financial metrics. Error metrics include Root Mean Squared Error (RMSE), Mean Average Percent Error (MAPE) and Continous Ranked Probability Score (CRPS) where lower for all three is preferred. RMSE and MAPE measure the error of the forecasts in a point forecast setting while CRPS measures error of the forecasted distribution. Accuracy metrics include the binary accuracy of the sign of the mean of the forecast. Although 50\% seems like a non-trivial baseline for the binary settings (i.e., price either goes up or down), the S\&P500 has a historical upwards trend. Instead, we consider the historical binary accuracy as a non-trivial baseline. Financial metrics include the returns of the trading strategies and the return of a passive trading strategy as a baseline. An in depth explanation of all three classes of metrics are described in Section~\ref{appendix:metrics} in Appendix.

\vspace*{-4mm}

\subsection{Validation of pipeline}
\vspace*{-1mm}
Results for both the training and testing splits are shown in Table~\ref{table:finance_table} for the 8 tickers in the daily setting (hourly setting shown in Table~\ref{table:finance_table_hourly} in Appendix). Accounting for the 8 tickers and both frequencies, we have 16 possibilities referred to as \textit{settings}. Across all 16 settings, the performance on training data is strong for both ML and financial metrics. On the testing split, performance is weaker but we still have that 6 out of the 8 tickers have at least 1 strategy with returns greater than that of the passive strategy. Hence, we believe that our pipeline is a good approximation (albeit a significantly simpler one) for a pipeline that could potentially be implemented  in industry for assisting intra-day trading. Section~\ref{appendix:pipeline_strength} in the Appendix discusses both tables in more depth.

\begin{table*}[t]
\large
\resizebox{\linewidth}{!}{
\begin{tabular}{llllllllll}
\toprule
\multicolumn{2}{c}{Setting} & \multicolumn{2}{c}{Accuracy} & \multicolumn{5}{c}{Returns} \\ 
\midrule
\multicolumn{1}{c}{Ticker} & \multicolumn{1}{c}{Set} &  \multicolumn{1}{c}{T} & \multicolumn{1}{c}{P} & \multicolumn{1}{c}{Passive} & \multicolumn{1}{c}{$\tau=0$} & \multicolumn{1}{c}{Kelly $\tau = 0$} & \multicolumn{1}{c}{$\tau = \mu_y + \sigma_y$} & \multicolumn{1}{c}{Kelly $\tau = \mu_y + \sigma_y$}  \\ 
 \toprule 
\multirow{2}{*}{\textbf{ADSK}} &\multicolumn{1}{c}{\textbf{train}} & \multicolumn{1}{c}{55.7}  & \multicolumn{1}{c}{70.4}  & \multicolumn{1}{c}{260.7}  & \multicolumn{1}{c}{\textbf{6148.1}}  & \multicolumn{1}{c}{\textbf{763.2}}  & \multicolumn{1}{c}{152.9}  & \multicolumn{1}{c}{102.6}  \\ 
 &\multicolumn{1}{c}{\textbf{test}} & \multicolumn{1}{c}{52.6}  & \multicolumn{1}{c}{46.3}  & \multicolumn{1}{c}{18.0}  & \multicolumn{1}{c}{-11.3}  & \multicolumn{1}{c}{-1.3}  & \multicolumn{1}{c}{2.2}  & \multicolumn{1}{c}{-1.4}  \\ 
\midrule
\multirow{2}{*}{\textbf{BWA}} &\multicolumn{1}{c}{\textbf{train}} & \multicolumn{1}{c}{52.1}  & \multicolumn{1}{c}{73.7}  & \multicolumn{1}{c}{-16.4}  & \multicolumn{1}{c}{\textbf{3850.2}}  & \multicolumn{1}{c}{\textbf{585.0}}  & \multicolumn{1}{c}{\textbf{326.1}}  & \multicolumn{1}{c}{\textbf{190.7}}  \\ 
 &\multicolumn{1}{c}{\textbf{test}} & \multicolumn{1}{c}{50.5}  & \multicolumn{1}{c}{60.0}  & \multicolumn{1}{c}{3.4}  & \multicolumn{1}{c}{\textbf{4.3}}  & \multicolumn{1}{c}{-0.4}  & \multicolumn{1}{c}{\textbf{7.4}}  & \multicolumn{1}{c}{-0.8}  \\ 
\midrule
\multirow{2}{*}{\textbf{CAH}} &\multicolumn{1}{c}{\textbf{train}} & \multicolumn{1}{c}{52.6}  & \multicolumn{1}{c}{56.3}  & \multicolumn{1}{c}{-36.3}  & \multicolumn{1}{c}{\textbf{179.2}}  & \multicolumn{1}{c}{\textbf{-2.5}}  & \multicolumn{1}{c}{\textbf{10.0}}  & \multicolumn{1}{c}{\textbf{1.3}}  \\ 
 &\multicolumn{1}{c}{\textbf{test}} & \multicolumn{1}{c}{55.8}  & \multicolumn{1}{c}{42.1}  & \multicolumn{1}{c}{-1.7}  & \multicolumn{1}{c}{-11.7}  & \multicolumn{1}{c}{\textbf{0.6}}  & \multicolumn{1}{c}{\textbf{8.3}}  & \multicolumn{1}{c}{\textbf{4.4}}  \\ 
\midrule
\multirow{2}{*}{\textbf{CE}} &\multicolumn{1}{c}{\textbf{train}} & \multicolumn{1}{c}{51.4}  & \multicolumn{1}{c}{73.7}  & \multicolumn{1}{c}{47.2}  & \multicolumn{1}{c}{\textbf{1809.8}}  & \multicolumn{1}{c}{\textbf{332.4}}  & \multicolumn{1}{c}{\textbf{165.1}}  & \multicolumn{1}{c}{\textbf{82.2}}  \\ 
 &\multicolumn{1}{c}{\textbf{test}} & \multicolumn{1}{c}{52.6}  & \multicolumn{1}{c}{66.3}  & \multicolumn{1}{c}{33.7}  & \multicolumn{1}{c}{\textbf{42.8}}  & \multicolumn{1}{c}{21.3}  & \multicolumn{1}{c}{5.9}  & \multicolumn{1}{c}{5.0}  \\ 
\midrule
\multirow{2}{*}{\textbf{CHTR}} &\multicolumn{1}{c}{\textbf{train}} & \multicolumn{1}{c}{53.0}  & \multicolumn{1}{c}{80.6}  & \multicolumn{1}{c}{205.3}  & \multicolumn{1}{c}{\textbf{12382.9}}  & \multicolumn{1}{c}{\textbf{3168.6}}  & \multicolumn{1}{c}{\textbf{608.7}}  & \multicolumn{1}{c}{\textbf{355.5}}  \\ 
 &\multicolumn{1}{c}{\textbf{test}} & \multicolumn{1}{c}{54.7}  & \multicolumn{1}{c}{55.8}  & \multicolumn{1}{c}{10.7}  & \multicolumn{1}{c}{\textbf{12.2}}  & \multicolumn{1}{c}{-0.5}  & \multicolumn{1}{c}{5.1}  & \multicolumn{1}{c}{-0.1}  \\ 
\midrule
\multirow{2}{*}{\textbf{HAS}} &\multicolumn{1}{c}{\textbf{train}} & \multicolumn{1}{c}{50.0}  & \multicolumn{1}{c}{63.3}  & \multicolumn{1}{c}{7.2}  & \multicolumn{1}{c}{\textbf{661.6}}  & \multicolumn{1}{c}{-1.1}  & \multicolumn{1}{c}{\textbf{80.8}}  & \multicolumn{1}{c}{\textbf{23.6}}  \\ 
 &\multicolumn{1}{c}{\textbf{test}} & \multicolumn{1}{c}{50.0}  & \multicolumn{1}{c}{58.3}  & \multicolumn{1}{c}{20.9}  & \multicolumn{1}{c}{19.2}  & \multicolumn{1}{c}{-0.3}  & \multicolumn{1}{c}{11.8}  & \multicolumn{1}{c}{6.4}  \\ 
\midrule
\multirow{2}{*}{\textbf{FANG}} &\multicolumn{1}{c}{\textbf{train}} & \multicolumn{1}{c}{50.7}  & \multicolumn{1}{c}{69.6}  & \multicolumn{1}{c}{-29.5}  & \multicolumn{1}{c}{\textbf{5728.9}}  & \multicolumn{1}{c}{\textbf{383.8}}  & \multicolumn{1}{c}{\textbf{418.7}}  & \multicolumn{1}{c}{\textbf{247.4}}  \\ 
 &\multicolumn{1}{c}{\textbf{test}} & \multicolumn{1}{c}{54.7}  & \multicolumn{1}{c}{58.9}  & \multicolumn{1}{c}{26.7}  & \multicolumn{1}{c}{\textbf{43.2}}  & \multicolumn{1}{c}{5.3}  & \multicolumn{1}{c}{0.8}  & \multicolumn{1}{c}{-6.1}  \\ 
\midrule
\multirow{2}{*}{\textbf{XLNX}} &\multicolumn{1}{c}{\textbf{train}} & \multicolumn{1}{c}{52.3}  & \multicolumn{1}{c}{54.6}  & \multicolumn{1}{c}{104.1}  & \multicolumn{1}{c}{\textbf{104.9}}  & \multicolumn{1}{c}{-4.7}  & \multicolumn{1}{c}{15.7}  & \multicolumn{1}{c}{2.3}  \\ 
 &\multicolumn{1}{c}{\textbf{test}} & \multicolumn{1}{c}{52.6}  & \multicolumn{1}{c}{54.7}  & \multicolumn{1}{c}{34.8}  & \multicolumn{1}{c}{11.5}  & \multicolumn{1}{c}{-0.7}  & \multicolumn{1}{c}{1.7}  & \multicolumn{1}{c}{2.3}  \\ 
\midrule
\end{tabular}

}
\caption{Performance of pipeline for the 8 tickers in the daily setting. Error metrics are omitted as they are not all comparable across the training and testing splits. Strategies with returns above that of the passive strategy are bolded.}
\label{table:finance_table}
\end{table*}

\vspace*{-4mm}
\section{Evaluation Results}
\vspace*{-2mm}
We now apply our interpretability method to our pipeline and evaluate the resulting interpretability benefits. In Section~\ref{ssec:adv_example_results}, we investigate the  distribution parameters ($\mu$, $\sigma$ and $\nu$) forecasted by our model in the presence of manipulations introduced by our algorithm. In Section~\ref{ssec:returns}, we analyze these changes with the entire pipeline in mind, that is we measure the performance of trading strategies in the synthetic market setting generated by our algorithm. We see how hyperparameters of Algorithm~\ref{algo1} provide controllability on the adversity of the synthetic setting, allowing us to draw insights into the forecasting pipeline's modeling. In particular,  we analyze the importance of different features in explaining the pipeline's performance in Section~\ref{ssec:changes_in_features}. %

\vspace*{-4mm}
\subsection{Interpretability of the Forecasting Model}
\vspace*{-2mm}
\label{ssec:adv_example_results}

\begin{table*}[t]
\large
\resizebox{\linewidth}{!}{
\begin{tabular}{llllllllllllll}
\toprule
\multicolumn{4}{c}{Setting} & \multicolumn{3}{c}{Error} & \multicolumn{2}{c}{Accuracy} & \multicolumn{5}{c}{Returns} \\ 
\midrule
\multicolumn{1}{c}{Ticker} & \multicolumn{1}{c}{Set} & \multicolumn{1}{c}{Parameter} & \multicolumn{1}{c}{Direction} & \multicolumn{1}{c}{RMSE} & \multicolumn{1}{c}{MAPE} & \multicolumn{1}{c}{CRSP} & \multicolumn{1}{c}{T} & \multicolumn{1}{c}{P} & \multicolumn{1}{c}{Passive} & \multicolumn{1}{c}{$\tau = 0$} & \multicolumn{1}{c}{Kelly $\tau = 0$} & \multicolumn{1}{c}{$\tau = \mu_y +\sigma_y$} & \multicolumn{1}{c}{Kelly $\tau = \mu_y +\sigma_y$}  \\ 
 \midrule 
\multirow{8}{*}{\textbf{ADSK}} &\multicolumn{1}{c}{\textbf{Testing}} & \multicolumn{1}{c}{\NA}  & \multicolumn{1}{c}{\NA}  & \multicolumn{1}{c}{0.01985}  & \multicolumn{1}{c}{1.67056}  & \multicolumn{1}{c}{0.91933}  & \multirow{7}{*}{52.6}  & \multicolumn{1}{c}{46.3}  & \multicolumn{1}{c}{18.0}  & \multicolumn{1}{c}{-11.3}  & \multicolumn{1}{c}{-1.3}  & \multicolumn{1}{c}{2.2}  & \multicolumn{1}{c}{-1.4}  \\ 
\cline{2-14}
 & \multirow{6}{*}{\textbf{Synthetic}}  & \multicolumn{1}{c}{$\mu$}  & \multicolumn{1}{c}{$\uparrow$}  & \multicolumn{1}{c}{\cellcolor{green!25}0.02382}  & \multicolumn{1}{c}{\cellcolor{green!25}4.1041}  & \multicolumn{1}{c}{\cellcolor{green!25}1.09991}  &  & \multicolumn{1}{c}{49.5}  & \multicolumn{1}{c}{18.0}  & \multicolumn{1}{c}{-8.5}  & \multicolumn{1}{c}{-0.4}  & \multicolumn{1}{c}{\cellcolor{green!25}-6.6}  & \multicolumn{1}{c}{-0.4}  \\ 
 &    & \multicolumn{1}{c}{$\mu$}  & \multicolumn{1}{c}{$\downarrow$}  & \multicolumn{1}{c}{\cellcolor{green!25}0.02079}  & \multicolumn{1}{c}{\cellcolor{green!25}1.97913}  & \multicolumn{1}{c}{\cellcolor{green!25}0.95181}  &  & \multicolumn{1}{c}{48.4}  & \multicolumn{1}{c}{18.0}  & \multicolumn{1}{c}{-8.1}  & \multicolumn{1}{c}{-0.1}  & \multicolumn{1}{c}{\cellcolor{green!25}-4.9}  & \multicolumn{1}{c}{-0.4}  \\ 
 &    & \multicolumn{1}{c}{$\sigma$}  & \multicolumn{1}{c}{$\uparrow$}  & \multicolumn{1}{c}{\cellcolor{green!25}0.02009}  & \multicolumn{1}{c}{\cellcolor{green!25}2.13263}  & \multicolumn{1}{c}{\cellcolor{green!25}0.95911}  &  & \multicolumn{1}{c}{48.4}  & \multicolumn{1}{c}{18.0}  & \multicolumn{1}{c}{-6.9}  & \multicolumn{1}{c}{\cellcolor{green!25}-1.9}  & \multicolumn{1}{c}{\cellcolor{green!25}-2.3}  & \multicolumn{1}{c}{-0.4}  \\ 
 &    & \multicolumn{1}{c}{$\sigma$}  & \multicolumn{1}{c}{$\downarrow$}  & \multicolumn{1}{c}{\cellcolor{green!25}0.02047}  & \multicolumn{1}{c}{\cellcolor{green!25}2.09409}  & \multicolumn{1}{c}{\cellcolor{green!25}0.96214}  &  & \multicolumn{1}{c}{50.5}  & \multicolumn{1}{c}{18.0}  & \multicolumn{1}{c}{\cellcolor{green!25}-11.8}  & \multicolumn{1}{c}{-0.4}  & \multicolumn{1}{c}{10.3}  & \multicolumn{1}{c}{-1.4}  \\ 
 &    & \multicolumn{1}{c}{$\nu$}  & \multicolumn{1}{c}{$\uparrow$}  & \multicolumn{1}{c}{\cellcolor{green!25}0.02103}  & \multicolumn{1}{c}{\cellcolor{green!25}2.2685}  & \multicolumn{1}{c}{\cellcolor{green!25}1.01209}  &  & \multicolumn{1}{c}{\cellcolor{green!25}42.1}  & \multicolumn{1}{c}{18.0}  & \multicolumn{1}{c}{\cellcolor{green!25}-16.8}  & \multicolumn{1}{c}{\cellcolor{green!25}-2.3}  & \multicolumn{1}{c}{\cellcolor{green!25}-3.5}  & \multicolumn{1}{c}{-0.4}  \\ 
 &    & \multicolumn{1}{c}{$\nu$}  & \multicolumn{1}{c}{$\downarrow$}  & \multicolumn{1}{c}{\cellcolor{green!25}0.0207}  & \multicolumn{1}{c}{\cellcolor{green!25}2.43082}  & \multicolumn{1}{c}{\cellcolor{green!25}1.01386}  &  & \multicolumn{1}{c}{52.6}  & \multicolumn{1}{c}{18.0}  & \multicolumn{1}{c}{\cellcolor{green!25}-12.4}  & \multicolumn{1}{c}{-0.4}  & \multicolumn{1}{c}{3.1}  & \multicolumn{1}{c}{-0.4}  \\ 
\bottomrule
\multirow{8}{*}{\textbf{BWA}} &\multicolumn{1}{c}{\textbf{Testing}} & \multicolumn{1}{c}{\NA}  & \multicolumn{1}{c}{\NA}  & \multicolumn{1}{c}{0.0191}  & \multicolumn{1}{c}{1.34892}  & \multicolumn{1}{c}{0.86445}  & \multirow{7}{*}{50.5}  & \multicolumn{1}{c}{60.0}  & \multicolumn{1}{c}{3.4}  & \multicolumn{1}{c}{4.3}  & \multicolumn{1}{c}{-0.4}  & \multicolumn{1}{c}{7.4}  & \multicolumn{1}{c}{-0.8}  \\ 
\cline{2-14}
 & \multirow{6}{*}{\textbf{Synthetic}}  & \multicolumn{1}{c}{$\mu$}  & \multicolumn{1}{c}{$\uparrow$}  & \multicolumn{1}{c}{\cellcolor{green!25}0.02102}  & \multicolumn{1}{c}{\cellcolor{green!25}1.95462}  & \multicolumn{1}{c}{\cellcolor{green!25}0.99858}  &  & \multicolumn{1}{c}{\cellcolor{green!25}53.7}  & \multicolumn{1}{c}{3.4}  & \multicolumn{1}{c}{\cellcolor{green!25}-4.6}  & \multicolumn{1}{c}{\cellcolor{green!25}-0.4}  & \multicolumn{1}{c}{\cellcolor{green!25}4.2}  & \multicolumn{1}{c}{-0.1}  \\ 
 &    & \multicolumn{1}{c}{$\mu$}  & \multicolumn{1}{c}{$\downarrow$}  & \multicolumn{1}{c}{\cellcolor{green!25}0.01965}  & \multicolumn{1}{c}{\cellcolor{green!25}1.61631}  & \multicolumn{1}{c}{\cellcolor{green!25}0.90919}  &  & \multicolumn{1}{c}{\cellcolor{green!25}53.7}  & \multicolumn{1}{c}{3.4}  & \multicolumn{1}{c}{\cellcolor{green!25}-4.1}  & \multicolumn{1}{c}{1.3}  & \multicolumn{1}{c}{\cellcolor{green!25}6.3}  & \multicolumn{1}{c}{-0.1}  \\ 
 &    & \multicolumn{1}{c}{$\sigma$}  & \multicolumn{1}{c}{$\uparrow$}  & \multicolumn{1}{c}{\cellcolor{green!25}0.01921}  & \multicolumn{1}{c}{\cellcolor{green!25}1.69147}  & \multicolumn{1}{c}{\cellcolor{green!25}0.88259}  &  & \multicolumn{1}{c}{62.1}  & \multicolumn{1}{c}{3.4}  & \multicolumn{1}{c}{8.7}  & \multicolumn{1}{c}{-0.4}  & \multicolumn{1}{c}{9.8}  & \multicolumn{1}{c}{5.2}  \\ 
 &    & \multicolumn{1}{c}{$\sigma$}  & \multicolumn{1}{c}{$\downarrow$}  & \multicolumn{1}{c}{\cellcolor{green!25}0.01947}  & \multicolumn{1}{c}{\cellcolor{green!25}1.52496}  & \multicolumn{1}{c}{\cellcolor{green!25}0.88966}  &  & \multicolumn{1}{c}{\cellcolor{green!25}54.7}  & \multicolumn{1}{c}{3.4}  & \multicolumn{1}{c}{\cellcolor{green!25}-9.6}  & \multicolumn{1}{c}{\cellcolor{green!25}-0.4}  & \multicolumn{1}{c}{\cellcolor{green!25}6.4}  & \multicolumn{1}{c}{-0.1}  \\ 
 &    & \multicolumn{1}{c}{$\nu$}  & \multicolumn{1}{c}{$\uparrow$}  & \multicolumn{1}{c}{0.01901}  & \multicolumn{1}{c}{\cellcolor{green!25}1.62888}  & \multicolumn{1}{c}{\cellcolor{green!25}0.86666}  &  & \multicolumn{1}{c}{\cellcolor{green!25}54.7}  & \multicolumn{1}{c}{3.4}  & \multicolumn{1}{c}{\cellcolor{green!25}-3.5}  & \multicolumn{1}{c}{-0.1}  & \multicolumn{1}{c}{\cellcolor{green!25}3.0}  & \multicolumn{1}{c}{\cellcolor{green!25}-1.0}  \\ 
 &    & \multicolumn{1}{c}{$\nu$}  & \multicolumn{1}{c}{$\downarrow$}  & \multicolumn{1}{c}{\cellcolor{green!25}0.02233}  & \multicolumn{1}{c}{\cellcolor{green!25}2.65821}  & \multicolumn{1}{c}{\cellcolor{green!25}1.05353}  &  & \multicolumn{1}{c}{\cellcolor{green!25}53.7}  & \multicolumn{1}{c}{3.4}  & \multicolumn{1}{c}{\cellcolor{green!25}-6.0}  & \multicolumn{1}{c}{\cellcolor{green!25}-0.4}  & \multicolumn{1}{c}{\cellcolor{green!25}5.4}  & \multicolumn{1}{c}{-0.1}  \\ 
\bottomrule
\end{tabular}
}
\caption{Performance of pipeline on synthetic features when varying all three distribution parameters in both directions for \$ADSK and \$BWA in the daily frequency. Performance on testing distribution shown for comparison. Metrics that degraded in the synthetic setting are in green.}
\label{adv_table}
\end{table*}

In Table~\ref{adv_table}, we show the  performance degradation resulting from manipulating the student-T distribution parameters in both directions for \$ADSK and \$BWA for daily trading. We include the testing performance for comparison. Performance for the remaining 6 tickers and the hourly setting are shown in Table~\ref{big_adv_table} and~\ref{table:deepar-hourly} in the Appendix. Settings where performance on $\hat{\mathbf{X}}$ is worse than that of $\mathbf{X}$ are colored in green.
In Table~\ref{adv_table}, all error metrics (RMSE, MAPE, CRSP) have their performance drop in all 6 possible synthetic setting. 

\begin{figure}[t]
    \centering
    \includegraphics[width =0.8\columnwidth]{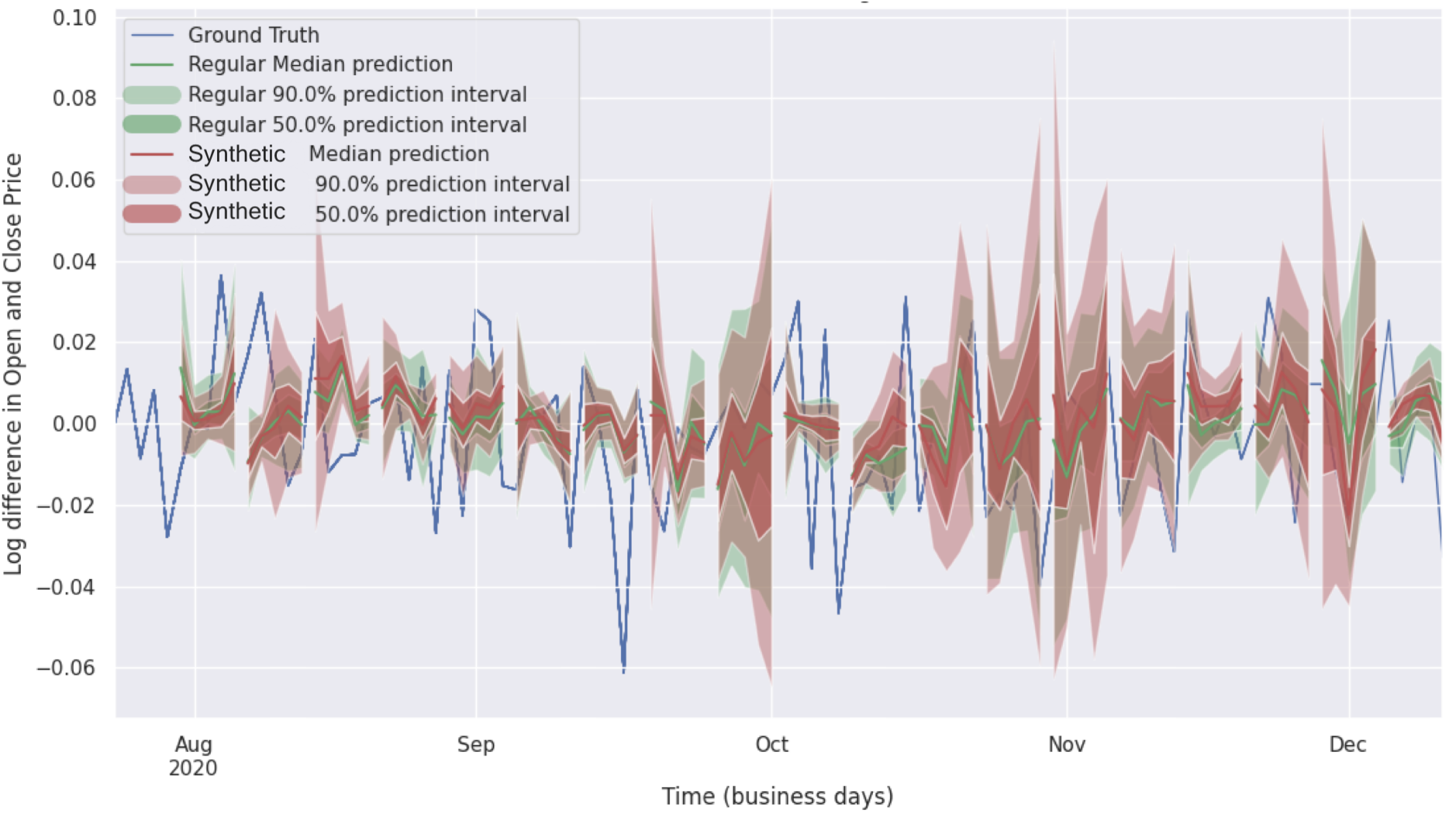}
    \caption{Regular (green) and synthetic (red) forecast for increasing the standard deviation $\sigma$ of \$BWA. Ground truth is shown in blue. The confidence intervals of the synthetic setting overshadows that of the regular setting.}
    \label{fig:increase sigma}
\end{figure}

\begin{figure}[t]
    \centering
    \includegraphics[width = 0.8\columnwidth]{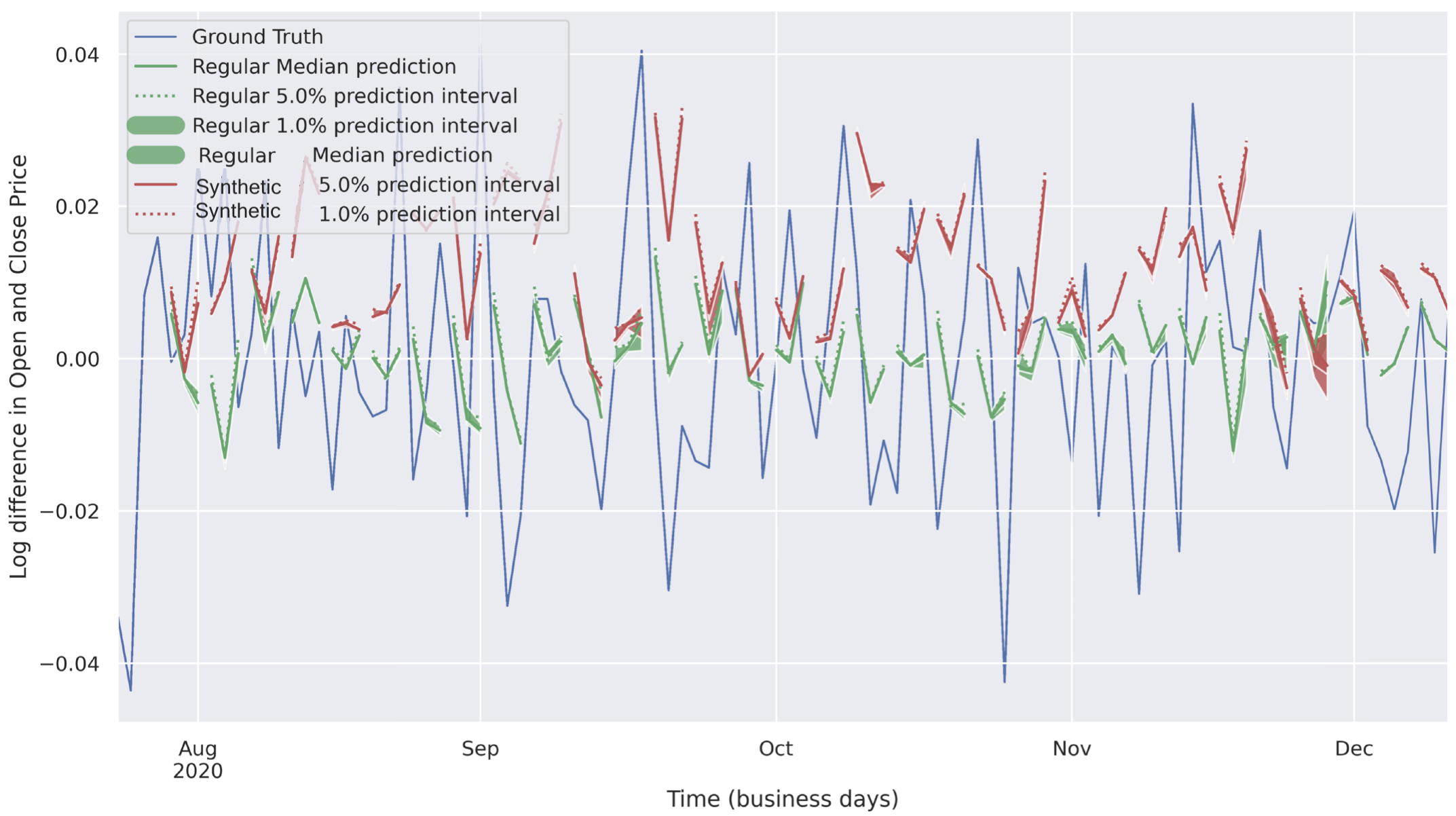}
    \caption{Regular (green) and synthetic (red) forecast for increasing the mean $\mu$ of \$HAS. Ground truth is shown in blue. The adversarial forecasts lie significantly above the regular forecasts.}
    \label{fig:increase mu}
\end{figure}

These drops in performance can be understood via Figure~\ref{fig:increase sigma} where we plot the resulting forecasts for increasing the standard deviation. The adversarial confidence interval (red) overshadows that of the benign setting (green), which degrades the CRSP for instance. The same is shown for increasing the mean in Figure~\ref{fig:increase mu} where the adversarial mean (red) lies above the regular mean (green) which degrades the RMSE and MAPE. When considering binary accuracy and financial returns, we see a mix of small and large improvements or degradations in the synthetic setting. For instance, the return for \$BWA kelly $\tau = 0$ strategy increases by 2\% in the synthetic setting when increasing $\mu$ in Table~\ref{adv_table} but decreases by almost 10\% for the non-kelly strategy. A similar trend is observed for all 8 companies and both daily and hourly frequency (in the Appendix).

\vspace*{-5mm}
\subsubsection{Performance of post-processing}
The change in performance of the several metrics provides intuition on the importance of considering an end-to-end pipeline and not an ML model in isolation. When calculating RMSE and MAPE, the distribution is collapsed to a point estimate by taking the mean of the distribution. Accordingly, as such metrics are directly related to the mean, any perturbation results into large change in such metrics. For instance, the RMSE for ADSK changes by 10.2\% on average when manipulating the mean. When we consider distribution parameters that are not related to such metrics, we do not see a similar change in performance (only 0.21\% when manipulating the standard deviation).
On the other hand, the return of a trading strategy is a function of several variables and potentially non-smooth such as the threshold strategies. For instance, if we originally have $\mu \geq \tau$ and we decrease $\mu$ to get $\hat{\mu}$, if $\hat{\mu} \geq \tau$, the trade still occurs. Even if $\hat{\mu} < \tau$ and the trade does not occur, if the return of the trade $y$ was negative, we improve the overall return. Lastly, as the correct value of a distribution parameter is not known, we can move a distribution parameter in the correct direction. %
These example explain why more complex post-processing steps can result into both positive and negative changes that are equally useful to analyze.

\vspace*{-3mm}
\subsection{Looking into the Trading Strategies}
\label{ssec:returns}
\vspace*{-1mm}
We now investigate the regular and synthetic settings at the trading level by look at the return distributions of performed trades. The distribution of returns is a standard method used to diagnose the returns of a portfolio and where the majority of profit or loss occurs.

\vspace*{-5mm}
\subsubsection{Parameter Distributions and Direction.}

\begin{figure}[t]
  \subcaptionbox{$\tau = 0$ strategy. Regular setting has a mean return of 0.32\% and 40\% days traded. Synthetic setting has mean 0.17\% and 74.7\% days traded.\label{fig:6a}}%
  {\includegraphics[width=0.48\linewidth]{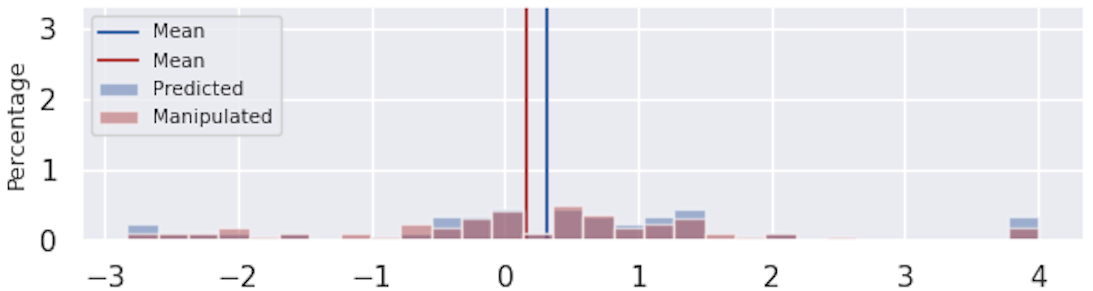}}
  \hspace{\fill}
  \subcaptionbox{$\tau = \mu_y + \sigma_y$ strategy. Regular setting has a mean return of 0.73\% and 7.4\% days traded. Synthetic setting has mean 0.27\% and 26.3\% days traded.\label{fig:6b}}%
  {\includegraphics[width=0.48\linewidth]{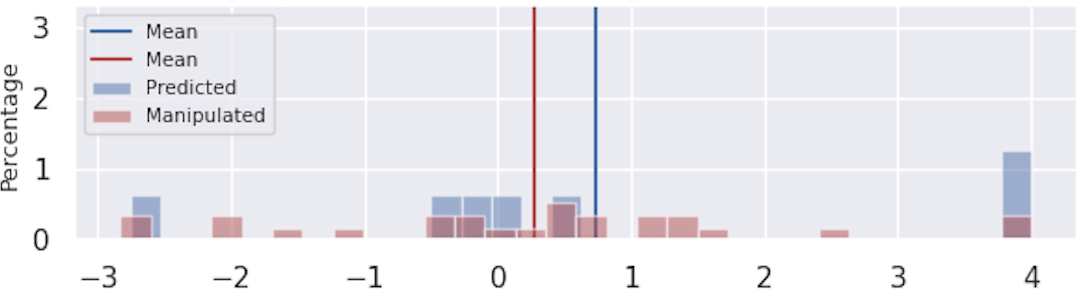}}

  \caption{Distribution of returns for both thresholding strategies on \$CHTR when increasing the mean.}\label{fig:increase mu testing}
\end{figure}

We first investigate how the choice of distribution parameters and directions of change may simulate different market setting to elicit varying effects on trading strategies.
Figure~\ref{fig:increase mu testing} shows the returns for the regular (blue) and synthetic (red) forecast when increasing the mean for both thresholding strategies. For $\tau = 0$ (subfigure~\ref{fig:6a}), we find that the number of days traded (i.e., days when $\mu \geq 0$) almost doubles (40.0\% to 74.7\%). However, new trades usually have a negative return which lower the average return. For a threshold of $\mu_y + \sigma_{y}$  (subfigure~\ref{fig:6b}), we see a similar but stronger effect as the number of trades almost quadruples and the total return thirds.

\vspace*{-5mm}
\subsubsection{Magnitude of Perturbation $\epsilon$.}
\begin{figure}[t]
  \subcaptionbox{$\tau = 0$ strategy for $\epsilon = \{0,0.01,0.03,0.1\}$ (left to right). The mean returns are 0.09\%, 0.05\%, 0.04\% and -0.23\% and percentage of days traded are 57.89\%, 57.89\%, 56.84\% and 68.42\%, from left to right.\label{fig:7a}}%
  {\includegraphics[width=\linewidth]{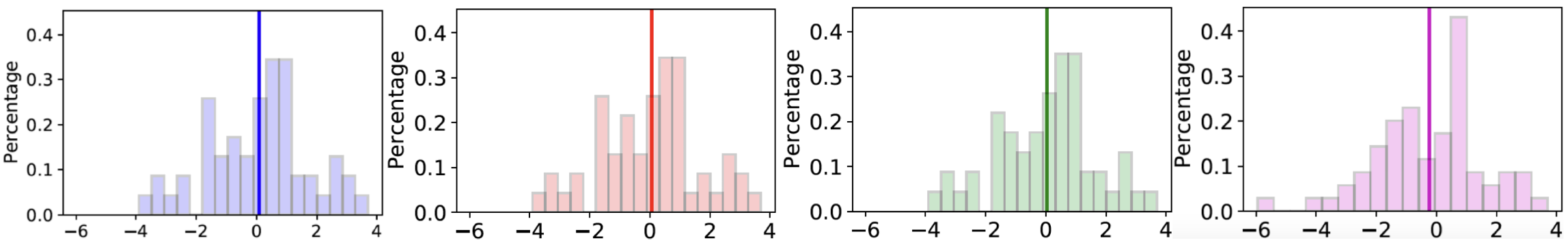}}
  \subcaptionbox{$\tau = \mu_y +\sigma_y$ strategy for $\epsilon = \{0,0.01,0.03,0.1\}$ (left to right). The mean returns are 0.8\%, 0.8\%, 0.49\% and 0.15\% and percentage of days traded are 9.47\%, 9.47\%, 10.53\% and 18.95\%, from left to right.\label{fig:7b}}%
  {\includegraphics[width=\linewidth]{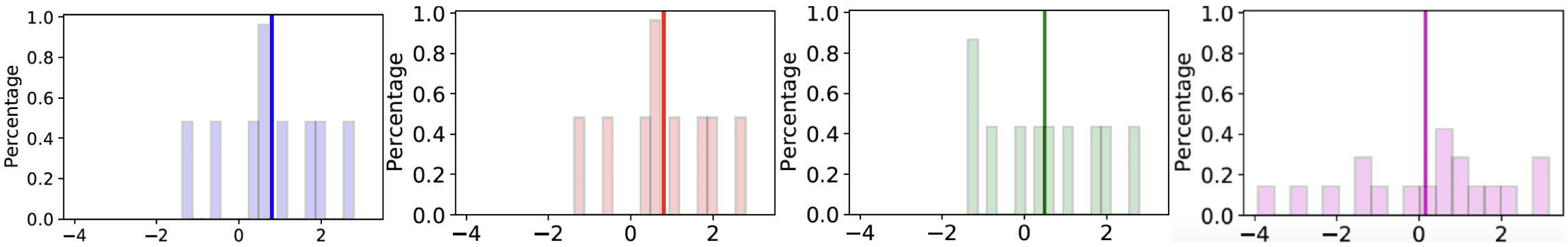}}
  \subcaptionbox{Fitting a Gaussian kernel on the above distributions for $\tau = 0$ (left) and $\tau = \mu_y + \sigma_y$ (right).\label{fig:7c}}%
  {\includegraphics[width=\linewidth]{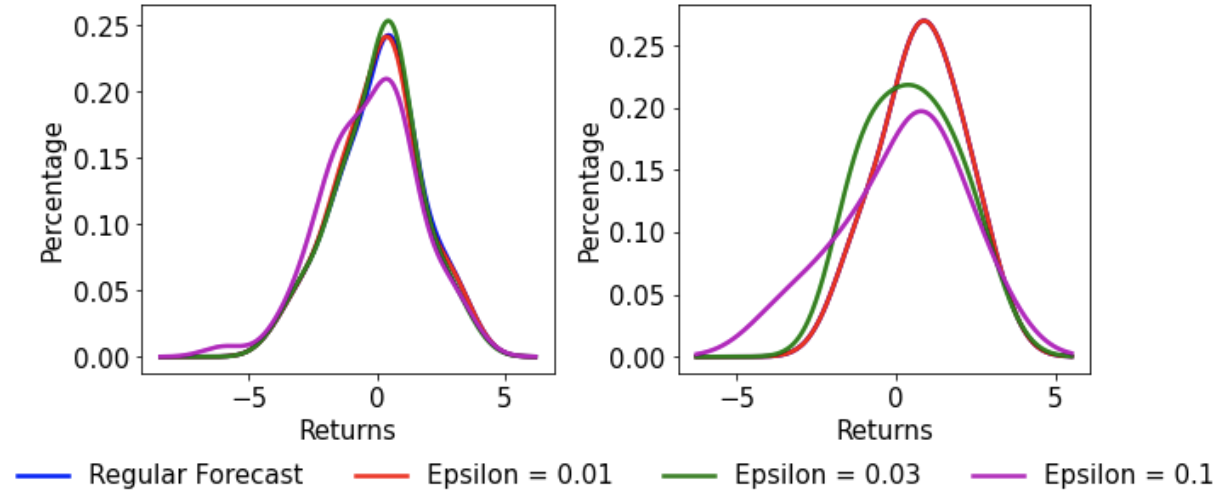}}
  \caption{Distribution of returns for \$BWA when increasing $\sigma$ for varying levels of epsilon. As $\epsilon$ increases (left to right), new (non-profitable) trades occur which decrease the mean return (vertical line).$\epsilon = 0$ represents the regular forecast.} 
  \label{fig:varying_eps}
\end{figure}

In Section~\ref{ssec:adv_example_results}, we fixed the perturbation size $\epsilon$. We now study our returns as this perturbation size grows. In Figure~\ref{fig:7a} and~\ref{fig:7b}, we report the returns distribution of both threshold strategies when increasing $\sigma$ for varying epsilons ($\epsilon = \{0.01,0.03,0.1\}$).  In Figure~\ref{fig:7c}, we fit a Gaussian kernel over each distribution to easily compare the distribution for varying $\epsilon$ values.

\setlength{\intextsep}{5pt}%
\setlength{\columnsep}{5pt}%
\begin{wrapfigure}{r}{0.45\textwidth}
    \includegraphics[width = 0.45\textwidth]{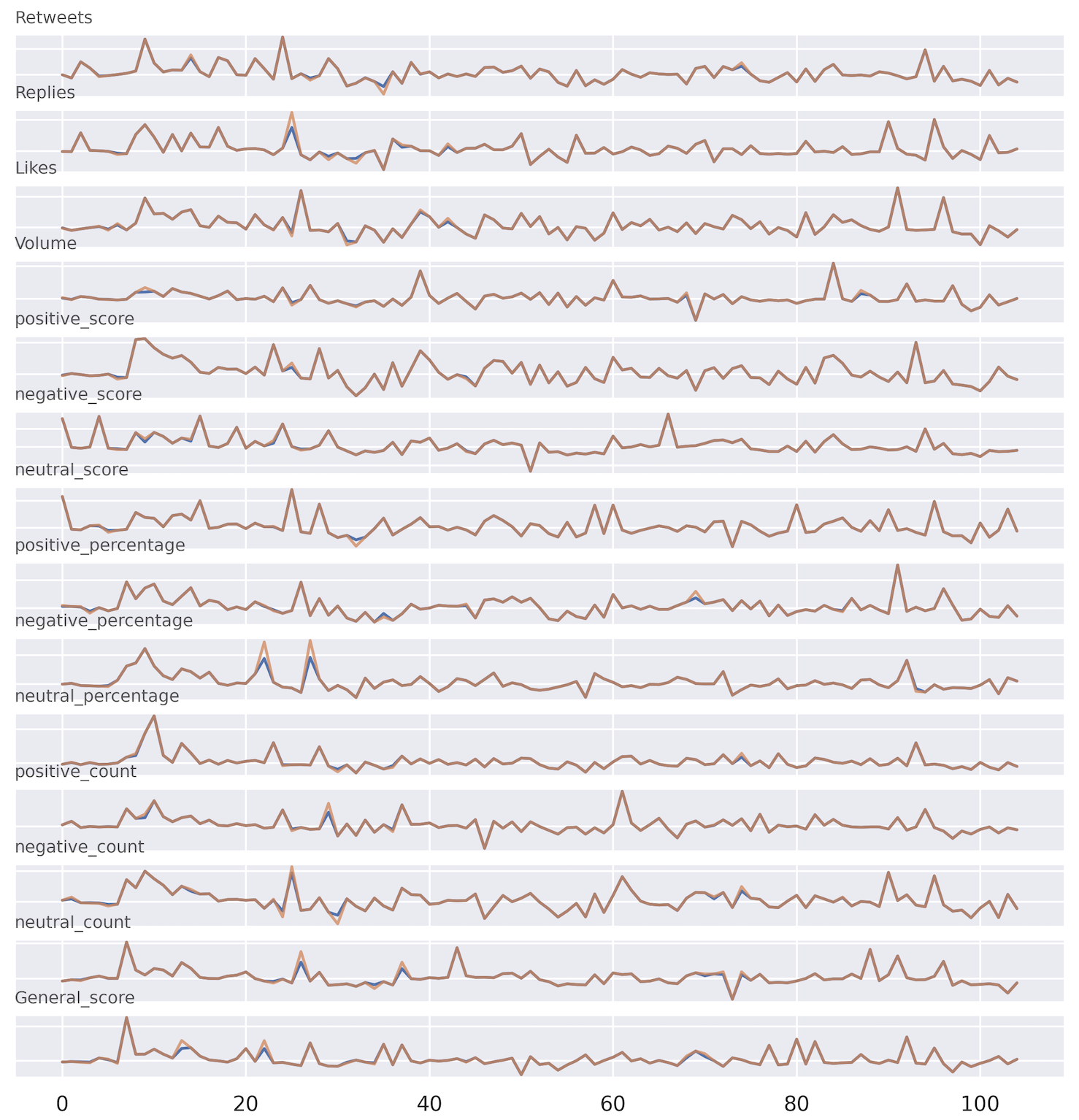}
    \caption{Regular (blue) and synthetic (orange) features for \$ADSK when increasing the mean $\mu$. Synthetic features are minimally different to regular features.}
    \label{fig:adv_features}
\end{wrapfigure}

We see that the mean of the distributions are shifting to the left as $\epsilon$ increases for both strategies in Figure~\ref{fig:varying_eps}. In addition, we are able to see larger negative tails as $\epsilon$ increases, especially for the $\tau =\mu_y + \sigma_y$ strategy in Figure~\ref{fig:7c}. As epsilon increases the percentage of negative returns increase. This is  clear for $\epsilon = 0.1$ (pink) for $\tau =\mu_y + \sigma_y$ strategy in Figure~\ref{fig:7b} where there are multiple trades with returns between -2\% to -4\% that do not occur in the regular setting (blue). Thus, the strength of the perturbation $\epsilon$ offers an intuitive and calibrated knob to simulate the level of adversity of the synthetic setting.
\vspace{-5mm}

\subsection{Perturbations at the feature level}
\label{ssec:changes_in_features}
\vspace*{-1mm}
We now investigate the perturbations at the feature level to understand their relative importance. An example of the regular and synthesized features are shown in Figure~\ref{fig:adv_features} ($\mathbf{X}$ in blue, $\hat{\mathbf{X}}$ in orange). We are able to determine which features are most important for a certain distribution parameter. Although the specific perturbation mask varied from in all settings, we find across several companies that increasing the standard deviation resulted into a significant perturbation in the positive sentiment features and decreasing the standard deviation was associated with negative sentiment features. Depending on the perturbation size $\epsilon$, the synthetic setting can reflect varying adversity including anomalous behaviour.

\subsection{Findings from our interpretability method}
\label{ssec:eval_conclusion}
\vspace{-2mm}
We showed how specific distribution parameters affect downstream calculations such as simple error metrics and trading strategies differently. Section~\ref{ssec:returns} demonstrates that the total perturbation amount in Algorithm~\ref{algo1} provides more control on how to evaluate the performance of a ML forecasting model and the entire pipeline end-to-end by controlling the adversity of the synthesized data. 
In our experiments, we found that small perturbations that manipulate the mean resulted into many non-profitable trades occurring despite these trades not occurring in the regular setting. Often, this resulted in large performance drops and financial returns. We encourage model owners to develop more robust models and pipelines able to more gracefully degrade their performance as the adversity of the synthetic setting increases.

\vspace{-3mm}
\subsubsection*{Acknowledgements}
This work was supported by CIFAR (through a Canada CIFAR AI Chair), by NSERC (under the Discovery Program, and COHESA strategic research network), and by a gift from Intel. We also thank the Vector Institute's sponsors.
\clearpage\section{Appendix}

\subsection{Forecasting Model Architecture}
\label{app:model_arch}
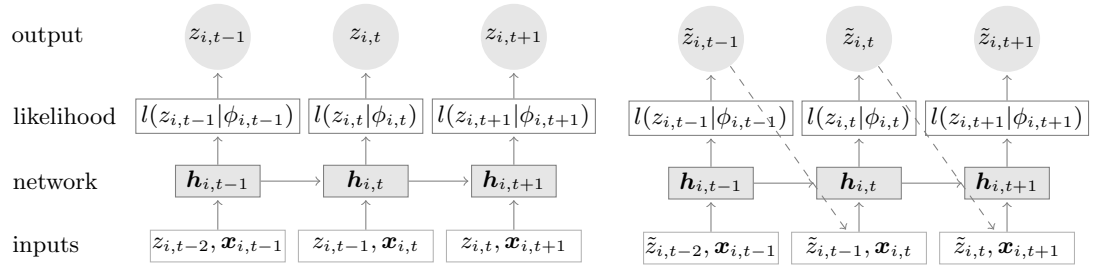
\begin{figure*}
    \centering
    \begin{tikzpicture}[shorten >=1pt,->,draw=black!50, font=\small, scale=0.48]%
    \tikzstyle{every pin edge}=[<-,shorten <=1pt]
    \tikzstyle{node}=[circle,fill=gray!20,minimum size=25pt,inner sep=0pt]
    \tikzstyle{box}=[rectangle,draw=black!50,minimum height=12pt, inner sep=0pt]
    \tikzstyle{input node}=[box, minimum width=51pt,draw=black!30];
    \tikzstyle{network node}=[box, draw, fill=gray!20, minimum width=32pt];
    \tikzstyle{output node}=[box, minimum width=42pt,inner sep=2pt];
    \tikzstyle{sample node}=[node];
    \tikzstyle{node dots}=[node, fill=black, scale=0.2];
    \tikzstyle{annot} = [text width=4em, text centered]
     \foreach \i/\t/\tm/\tp in {0/{t-1}/{t-2}/{t},1/{t}/{t-1}/{t+1},2/{t+1}/{t}/{t+2}} {%
         \node[input node] (x\i) at (4.1*\i,0 ) {$z_{i,{\tm}}, \bm{x}_{i, {\t}}$};%
       \node[network node] (f\i) at (4.1*\i,1.8) {$\bm{h}_{i, {\t}}$};%
       \node[output node] (y\i) at (4.1*\i,3.6) {$l(z_{i, \t}|\phi_{i, \t})$};%
         \node[sample node] (z\i) at (4.1*\i,5.8) {$z_{i, \t}$};
       \path (x\i) edge (f\i);%
       \path (f\i) edge (y\i);%
       \path (y\i) edge (z\i);%
     }%
     \node [left of=x0, node distance=0.90in, align=flush right] (xtlabel) {inputs};%
     \node [left of=f0, node distance=0.85in, align=flush right] (ftlabel) {network};%
     \node [left of=y0, node distance=0.8in, align=flush right] (ytlabel) {likelihood};
     \node [left of=z0, node distance=0.89in, align=flush right] (ztlabel) {output};
     \draw (f0) edge (f1);%
     \draw (f1) edge (f2);%
\end{tikzpicture}~~~~~~\begin{tikzpicture}[shorten >=1pt,->,draw=black!50, font=\small, scale=0.48]%
    \tikzstyle{every pin edge}=[<-,shorten <=1pt]
    \tikzstyle{node}=[circle,fill=gray!20,minimum size=25pt,inner sep=0pt]
    \tikzstyle{box}=[rectangle,draw=black!50,minimum height=12pt, inner sep=0pt]
    \tikzstyle{input node}=[box, minimum width=51pt,draw=black!30];
    \tikzstyle{network node}=[box, draw, fill=gray!20, minimum width=32pt];
    \tikzstyle{output node}=[box, minimum width=42pt,inner sep=2pt];
    \tikzstyle{sample node}=[node];
    \tikzstyle{node dots}=[node, fill=black, scale=0.2];
    \tikzstyle{annot} = [text width=4em, text centered]

     \foreach \i/\t/\tm/\tp in {0/{t-1}/{t-2}/{t},1/{t}/{t-1}/{t+1},2/{t+1}/{t}/{t+2}} {%
         \node[input node] (x\i) at (4.1*\i,0 ) {$\tilde{z}_{i,{\tm}}, \bm{x}_{i, {\t}}$};%
       \node[network node] (f\i) at (4.1*\i,1.8) {$\bm{h}_{i, {\t}}$};%
       \node[output node] (y\i) at (4.1*\i,3.6) {$l(z_{i, \t}|\phi_{i, \t})$};%
         \node[sample node] (z\i) at (4.1*\i,5.8) {$\tilde{z}_{i, \t}$};
       \path (x\i) edge (f\i);%
       \path (f\i) edge (y\i);%
       \path (y\i) edge (z\i);%
     }%
     \draw (f0) edge (f1);%
     \draw (f1) edge (f2);%
         \draw[dashed] (z0) edge (x1);%
         \draw[dashed] (z1) edge (x2);%
\end{tikzpicture}    \caption{Overview of the DeepAR model (figure taken from~\cite{salinas2020deepar}). Inputs $z_{i,t-1}$ and $\bm{x}_{i,t}$ as well as the previous RNN hidden state $\bm{h}_{i,t-1}$ are fed to the RNN's current state to compute $\bm{h}_{i,t}$ for each time step $t$. The RNN's output is then mapped to the parameters $\phi_{i,t}$ governing the likelihood function $l(z_{i,t}|\phi_{i, t})$ associated with a specific distributional assumption over $z_{i,t}$. training is depicted on the left for which we require $z_{i,t}$ to be known; autoregressive prediction is shown on the right where a sample $\tilde{z}_{i,t} \sim l(\cdot|\phi_{i, t})$ is drawn from the predictive distribution at $t$ and fed back into the prediction for $t+1$. }
    \label{fig:deepar_arch}
\end{figure*}

The model architecture of the DeepAR model is shown in Figure~\ref{fig:deepar_arch}. When using DeepAR we consider the log difference in open and close price as our single target time series and the Twitter features as covariates. In the forecasting setting, covariates are assumed to be known over the entire time period under consideration, including the prediction horizon of length $\tau$. This is not the case with Twitter data as we do not know future tweets. To alleviate this issue, we use lagged versions of the covariates such that no temporal violations occur.

\begin{table}
\centering
\begin{tabular}{c}
\toprule
Features \\
\midrule
Average positive score \\
Average negative score \\
Average neutral score \\
Percent \&number of positive tweets \\ 
Percent \&number of negative tweets \\ 
Percent \&number of neutral tweets \\ 
Average of each logit\\
Volume of tweets \\
Average likes\\
Average replies\\
\bottomrule
\end{tabular}
\caption{Sentiment and metadata based features used as inputs to the forecasting model.}
\label{table:sentiment features}
\end{table}

\subsection{Evaluation Details}
\label{appendix:metrics}
We provide details on our evaluation section below. We divide our historical data into a training, validation and testing split shown in figure~\ref{table:data_division}. 

\begin{table}
\centering
\begin{tabular}{r|ccc}
\multicolumn{1}{r}{Frequency}
& \multicolumn{1}{l}{Train}
& \multicolumn{1}{l}{Validation}
& \multicolumn{1}{l}{Test} \\ 
\midrule
Daily  & 0$-$1150 &1150$-$1190 & 1190$-$1306   \\
Hourly & 0$-$2550 &2550$-$2700 & 2700$-$2890
\end{tabular}
\caption{Division of data into the training, validation and testing sets in the daily and hourly settings.}
\label{table:data_division}
\end{table}

For our error metrics (RMSE, MAPE, CRPS) we leverage the ground truth values ($\mathbf{y}$) and our predictions ($\mathbf{\hat{y}}$) or the distributions themselves to determine the quality of the forecasts. The Root Mean Squared Error (RMSE) and Mean Average Percent Error are defined in Equation~\ref{eq:rmse} and~\ref{eq:mape}, respectively. They measure the residual error and percentage error between $\hat{y}$ and $y$, respectively. Smaller RMSE and MAPE are ideal. Both RMSE and MAPE are effectively point-estimate metrics as they only leverage the mean of the distribution. The Continous Ranked Probability Score (CRPS) measures the difference between the cumulative distribution function (CDF) of the forecast and the ground truth observation, shown in Equation~\ref{eq:crps}. A lower CRPS is preferred.

\begin{equation}
\label{eq:rmse}
\text{RMSE} = \sqrt{\frac{\sum_{t=1}^T (y_t - \hat{y}_t)^2 }{T}}
\end{equation}
\begin{equation}
\label{eq:mape}
\text{MAPE} = \frac{1}{T} \sum_{t=1}^T \left\lvert \frac{y_t - \hat{y}_t}{y_t} \right\rvert
\end{equation}
\begin{equation}
\label{eq:crps}
\text{CRPS}(F,x) = \int_{-\infty}^{\infty} \Big(F(y) -\mathbbm{1}(y-x)\Big)^{2} x^2 \,dy
\end{equation}

For our accuracy metrics, we consider binary accuracy. Binary accuracy refers to whether our point forecast $\mathbf{\hat{y}}$ have the same sign as the ground truth observation $\mathbf{y}$. In the binary setting, as the S\&P 500 tickers have had a historical upwards trend, comparing binary accuracy to 50\% is incorrect. Instead, we compare against the historical accuracy which is the maximum binary accuracy of a forecast that (1) always predicts up and (2) always predicts down. This is denoted by the \textbf{T} column in Table~\ref{table:finance_table} ("T" for "truth"). \textbf{P} in Table~\ref{table:finance_table} refers to the predicted binary accuracy from the forecasts.

For all trading strategies, we can calculate the net return of trading with such strategy. Similarly to binary accuracy, a return above 0\% is not a sufficient baseline for determining financial strength given the historical upwards trend of S\&P 500 tickers. Instead, we compare against the \emph{passive strategy baseline}: buying shares at the open price on the very first day of the split and selling it at the close price of the very last day. A \emph{non-trivial} strategy has a return above 0\% and also beats the passive strategy.

\subsection{Strength of trading pipeline}
\label{appendix:pipeline_strength}
We now provide a detailed discussion of the performance of the trading pipeline in the daily (Table~\ref{table:finance_table}) and hourly setting (Table~\ref{table:finance_table_hourly}). 

\paragraph{Strong performance on training split:} All 16 settings perform strongly across all metrics on the training split in Table~\ref{table:finance_table}. All 16 setting haves predicted binary accuracies above 50\% and 13 out of 16 beat the baseline binary accuracy, often by 10-15\%. For instance, daily trading of \$CHTR achieves a 80.6\% binary predicted accuracy, 27.6\% above the baseline of 53.0\%. All 16 settings have at least 1 of the 4 strategies beat the passive returns. For example, \$CHTR and and \$ADSK in the daily setting have a passive return of 205.3\% and 260.7\% in the training set while the $\tau = 0$ strategy beats them significantly at 12382.9\% and 6148.1\%, respectively.
\paragraph{(ii) Weaker (but positive) performance on testing split:} The performance drops significantly when looking at the testing set. In 7 out of 16 settings, the predicted binary accuracy beats the baseline predicted accuracy, often only by a couple percent. Even though the model is not trained for binary classification but rather the harder (probabilistic) regression task, performance on accuracy from point forecasts is biased. For the financial returns, all 16 settings have at least 1 of the 4 strategies have returns above 0\%. Additionally, 6 out of the 8 ticker have at least 1 strategy over both frequencies that outperforms the passive strategy. In total, 7 out of 16 settings have at least 1 strategy that outperforms the passive returns. On 1 extreme, \$ADSK and \$XLNX have 0 out of 8 settings that outperform the passive returns (albeit the passive returns are relatively high at 18\% and 27\% while the best strategy was still positive at 8.3\% and 8.4\%,respectively). On the other hand, \$BWA has 5 strategies out of 8 beat the passive returns (albeit the passive returns during that period were low at 3.4\% and 1.7\%, respectively).

\subsubsection{Are the forecasts meaningful?}
The weaker performance of the testing split compared to the training split is indicative of the difficulty of probabilistic regression tasks, especially of financial time series which are believed to be random walk models~\cite{random_walk}. Even though several best practices  were taken to avoid overfitting~\footnote{Dropout~\cite{dropout}, weight decay, early stopping and validation splits}, the results in Table~\ref{table:finance_table} attests to the difficulty of generalization in financial forecasting when backtesting on out-of-sample splits. If we compare to most other works on financial forecasting, criteria of success are often predicted binary accuracies above 50\% and returns above 0\% (if trading strategies are implemented at all)~\cite{bollen2011twitter,binary_acc}. Under such conditions, our pipeline can be considered non-trivial as out of the 16 settings and 4 strategies (ie: 64 possibilities), 45 have above 0\% returns. If we consider more stringent conditions (beating the baseline accuracy and passive trading strategy) that are more realistic, 6 out of the 8 tickers outperform the baseline accuracy and have at least 1 strategy that outperforms the passive strategy.

\begin{table*}[!h]
\large
\resizebox{\linewidth}{!}{
\begin{tabular}{llllllllll}
\toprule
\multicolumn{2}{c}{Setting} & \multicolumn{2}{c}{Accuracy} & \multicolumn{5}{c}{Returns} \\ 
\midrule
\multicolumn{1}{c}{Ticker} & \multicolumn{1}{c}{Set} &  \multicolumn{1}{c}{T} & \multicolumn{1}{c}{P} & \multicolumn{1}{c}{Passive} & \multicolumn{1}{c}{$\tau = 0$} & \multicolumn{1}{c}{Kelly $\tau = 0$} & \multicolumn{1}{c}{$\tau = \mu_y + \sigma_y$} & \multicolumn{1}{c}{Kelly $\tau = \mu_y + \sigma_y$}  \\ 
 \toprule 
\multirow{2}{*}{\textbf{ADSK}} &\multicolumn{1}{c}{\textbf{training}} & \multicolumn{1}{c}{57.4}  & \multicolumn{1}{c}{66.7}  & \multicolumn{1}{c}{40.0}  & \multicolumn{1}{c}{\textbf{1103.9}}  & \multicolumn{1}{c}{-0.1}  & \multicolumn{1}{c}{\textbf{107.7}}  & \multicolumn{1}{c}{\textbf{41.9}}  \\ 
 &\multicolumn{1}{c}{\textbf{test}} & \multicolumn{1}{c}{53.3}  & \multicolumn{1}{c}{51.8}  & \multicolumn{1}{c}{11.2}  & \multicolumn{1}{c}{8.3}  & \multicolumn{1}{c}{-0.1}  & \multicolumn{1}{c}{3.3}  & \multicolumn{1}{c}{1.1}  \\ 
\midrule
\multirow{2}{*}{\textbf{BWA}} &\multicolumn{1}{c}{\textbf{training}} & \multicolumn{1}{c}{55.4}  & \multicolumn{1}{c}{73.4}  & \multicolumn{1}{c}{1.2}  & \multicolumn{1}{c}{\textbf{2836.0}}  & \multicolumn{1}{c}{\textbf{352.7}}  & \multicolumn{1}{c}{\textbf{309.9}}  & \multicolumn{1}{c}{\textbf{141.4}}  \\ 
 &\multicolumn{1}{c}{\textbf{test}} & \multicolumn{1}{c}{56.4}  & \multicolumn{1}{c}{54.4}  & \multicolumn{1}{c}{1.7}  & \multicolumn{1}{c}{\textbf{10.9}}  & \multicolumn{1}{c}{-0.0}  & \multicolumn{1}{c}{\textbf{5.2}}  & \multicolumn{1}{c}{\textbf{2.3}}  \\ 
\midrule
\multirow{2}{*}{\textbf{CAH}} &\multicolumn{1}{c}{\textbf{training}} & \multicolumn{1}{c}{55.6}  & \multicolumn{1}{c}{74.1}  & \multicolumn{1}{c}{9.7}  & \multicolumn{1}{c}{\textbf{2231.7}}  & \multicolumn{1}{c}{1.5}  & \multicolumn{1}{c}{\textbf{284.1}}  & \multicolumn{1}{c}{\textbf{109.4}}  \\
\bottomrule
\end{tabular}

}
\caption{Performance of pipeline for the 8 tickers in the hourly setting. Error metrics are omitted as they are not all comparable across the training and testing splits. Strategies with returns above that of the passive strategy are bolded.}
\label{table:finance_table_hourly}
\end{table*}

\begin{table*}[!h]
\large
\resizebox{\linewidth}{!}{
\begin{tabular}{llllllllllllll}
\toprule
\multicolumn{4}{c}{Setting} & \multicolumn{3}{c}{Error} & \multicolumn{2}{c}{Accuracy} & \multicolumn{5}{c}{Returns} \\ 
\midrule
\multicolumn{1}{c}{Ticker} & \multicolumn{1}{c}{Set} & \multicolumn{1}{c}{Parameter} & \multicolumn{1}{c}{Direction} & \multicolumn{1}{c}{RMSE} & \multicolumn{1}{c}{MAPE} & \multicolumn{1}{c}{CRSP} & \multicolumn{1}{c}{T} & \multicolumn{1}{c}{P} & \multicolumn{1}{c}{Passive} & \multicolumn{1}{c}{$\tau = 0$} & \multicolumn{1}{c}{Kelly $\tau = 0$} & \multicolumn{1}{c}{$\tau = \mu_y +\sigma_y$} & \multicolumn{1}{c}{Kelly $\tau = \mu_y +\sigma_y$}  \\ 
 \midrule 
\multirow{8}{*}{\textbf{ADSK}} &\multicolumn{1}{c}{\textbf{test}} & \multicolumn{1}{c}{\NA}  & \multicolumn{1}{c}{\NA}  & \multicolumn{1}{c}{0.01985}  & \multicolumn{1}{c}{1.67056}  & \multicolumn{1}{c}{0.91933}  & \multirow{7}{*}{52.6}  & \multicolumn{1}{c}{46.3}  & \multicolumn{1}{c}{18.0}  & \multicolumn{1}{c}{-11.3}  & \multicolumn{1}{c}{-1.3}  & \multicolumn{1}{c}{2.2}  & \multicolumn{1}{c}{-1.4}  \\ 
\cline{2-14}
 & \multirow{6}{*}{\textbf{synthetic}}  & \multicolumn{1}{c}{$\mu$}  & \multicolumn{1}{c}{$\uparrow$}  & \multicolumn{1}{c}{\cellcolor{green!25}0.02382}  & \multicolumn{1}{c}{\cellcolor{green!25}4.1041}  & \multicolumn{1}{c}{\cellcolor{green!25}1.09991}  &  & \multicolumn{1}{c}{49.5}  & \multicolumn{1}{c}{18.0}  & \multicolumn{1}{c}{-8.5}  & \multicolumn{1}{c}{-0.4}  & \multicolumn{1}{c}{\cellcolor{green!25}-6.6}  & \multicolumn{1}{c}{-0.4}  \\ 
 &    & \multicolumn{1}{c}{$\mu$}  & \multicolumn{1}{c}{$\downarrow$}  & \multicolumn{1}{c}{\cellcolor{green!25}0.02079}  & \multicolumn{1}{c}{\cellcolor{green!25}1.97913}  & \multicolumn{1}{c}{\cellcolor{green!25}0.95181}  &  & \multicolumn{1}{c}{48.4}  & \multicolumn{1}{c}{18.0}  & \multicolumn{1}{c}{-8.1}  & \multicolumn{1}{c}{-0.1}  & \multicolumn{1}{c}{\cellcolor{green!25}-4.9}  & \multicolumn{1}{c}{-0.4}  \\
\bottomrule
\end{tabular}
}
\caption{Results of manipulating the 3 parameters of  Student-T distribution ($\mu$, $\sigma$ and $\nu$) in both directions using Algorithm~\ref{algo1} for all 8 tickers for the DeepAR model on daily forecasts. The metrics on the testing distribution are shown for comparison. Perturbations that degraded a metric are shown in green.}
\label{big_adv_table}
\end{table*}

\begin{table*}[h]
\large
\resizebox{\linewidth}{!}{
\begin{tabular}{llllllllllllll}
\toprule
\multicolumn{4}{c}{Setting} & \multicolumn{3}{c}{Error} & \multicolumn{2}{c}{Accuracy} & \multicolumn{5}{c}{Returns} \\ 
\midrule
\multicolumn{1}{c}{Ticker} & \multicolumn{1}{c}{Set} & \multicolumn{1}{c}{Parameter} & \multicolumn{1}{c}{Direction} & \multicolumn{1}{c}{RMSE} & \multicolumn{1}{c}{MAPE} & \multicolumn{1}{c}{CRSP} & \multicolumn{1}{c}{T} & \multicolumn{1}{c}{P} & \multicolumn{1}{c}{Passive} & \multicolumn{1}{c}{$\tau =0$} & \multicolumn{1}{c}{Kelly $\tau =0$} & \multicolumn{1}{c}{$\tau =\mu_y +\sigma_y$} & \multicolumn{1}{c}{Kelly $\tau =\mu_y +\sigma_y$}  \\ 
 \midrule 
\multirow{8}{*}{\textbf{ADSK}} &\multicolumn{1}{c}{\textbf{training}} & \multicolumn{1}{c}{\NA}  & \multicolumn{1}{c}{\NA}  & \multicolumn{1}{c}{0.00731}  & \multicolumn{1}{c}{1.04171}  & \multicolumn{1}{c}{0.66758}  & \multicolumn{1}{c}{57.4}  & \multicolumn{1}{c}{66.7}  & \multicolumn{1}{c}{40.0}  & \multicolumn{1}{c}{1103.9}  & \multicolumn{1}{c}{685.0}  & \multicolumn{1}{c}{107.7}  & \multicolumn{1}{c}{83.1}  \\ 
\cline{2-14}
 & \multicolumn{1}{c}{\textbf{test}} & \multicolumn{1}{c}{\NA}  & \multicolumn{1}{c}{\NA}  & \multicolumn{1}{c}{0.0067}  & \multicolumn{1}{c}{1.31622}  & \multicolumn{1}{c}{0.85351}  & \multirow{7}{*}{53.3}  & \multicolumn{1}{c}{51.8}  & \multicolumn{1}{c}{11.2}  & \multicolumn{1}{c}{8.3}  & \multicolumn{1}{c}{1.9}  & \multicolumn{1}{c}{3.3}  & \multicolumn{1}{c}{1.8}  \\ 
\cline{2-14}
 & \multirow{6}{*}{\textbf{synthetic}}  & \multicolumn{1}{c}{$\mu$}  & \multicolumn{1}{c}{$\uparrow$}  & \multicolumn{1}{c}{\cellcolor{green!25}0.00673}  & \multicolumn{1}{c}{\cellcolor{green!25}1.34907}  & \multicolumn{1}{c}{\cellcolor{green!25}0.86056}  &  & \multicolumn{1}{c}{\cellcolor{green!25}51.3}  & \multicolumn{1}{c}{11.2}  & \multicolumn{1}{c}{\cellcolor{green!25}8.1}  & \multicolumn{1}{c}{\cellcolor{green!25}1.6}  & \multicolumn{1}{c}{\cellcolor{green!25}2.2}  & \multicolumn{1}{c}{\cellcolor{green!25}0.8}  \\
\bottomrule
\end{tabular}
}
\caption{DeepAR Hourly}
\label{table:deepar-hourly}
\end{table*}

\bibliographystyle{splncs04}
\bibliography{mybibliography}

\end{document}